%% file: main.tex
\newcommand{\squeezeup}{\vspace{-0.6cm}}
\newcommand{\denselist}{\itemsep -2pt\parsep=1pt\partopsep 0pt}
\newcommand{\bitem}{\begin{itemize}\denselist}
\newcommand{\eitem}{\end{itemize}}
\newcommand{\benum}{\begin{enumerate}\denselist}
\newcommand{\eenum}{\end{enumerate}}

 \documentclass[9pt,sigconf]{acmart}

\setcopyright{acmcopyright}

\acmYear{2020}
\acmConference{ACM Sensys}{2020}{Yokohama Japan}
\acmPrice{}

\usepackage{xcolor}
\usepackage{amsfonts}
\usepackage{pifont}
\newcommand{\cmark}{\ding{51}}%
\newcommand{\xmark}{\ding{55}}%

\usepackage{color}
\usepackage{graphics}
\usepackage{graphicx}
\usepackage{url}
\usepackage{listings}
\usepackage{multicol}
\usepackage{multirow}
\usepackage[scaled]{helvet}
\usepackage{rotating}
\usepackage{xspace}
\usepackage{algorithm}
\usepackage[noend]{algpseudocode}
\usepackage{comment}
\usepackage{enumitem}
\usepackage{amsmath}
\usepackage{mathrsfs}
\usepackage{cancel}
\usepackage{cleveref}
\usepackage{subcaption}
\usepackage[labelfont=bf,textfont=it]{caption}

\newcommand{\name}{Pointillism\xspace}

\newcommand{\pcname}{Cross Potential Point Clouds\xspace}
\newcommand{\network}{RP-net\xspace}
\newcommand{\datasize}{54K\xspace}
\newcommand{\edited}[1]{\textcolor{black}{#1}}

\copyrightyear{2020}
\acmYear{2020}
\setcopyright{acmcopyright}\acmConference[SenSys '20]{The 18th ACM Conference on Embedded Networked Sensor Systems}{November 16--19, 2020}{Virtual Event, Japan}
\acmBooktitle{The 18th ACM Conference on Embedded Networked Sensor Systems (SenSys '20), November 16--19, 2020, Virtual Event, Japan}
\acmPrice{15.00}
\acmDOI{10.1145/3384419.3430783}
\acmISBN{978-1-4503-7590-0/20/11}

\begin{document}

\title{\name: Accurate 3D Bounding Box Estimation with Multi-Radars}

\author{Kshitiz Bansal$^*$, Keshav Rungta$^*$, Siyuan Zhu$^*$ and Dinesh Bharadia$^*$}
\affiliation{\institution{$^*$University of California San Diego}}
\renewcommand{\shortauthors}{K. Bansal, K. Rungta, S. Zhu, D. Bharadia}


\input{0-abstract}

\maketitle

\input{1-intro-v8}

\input{2-challenges.tex}


\input{3-singleVsMultiple-v2.tex}

\input{4-design-pcpotential-v2.tex}

\input{5-design-LLPnet.tex}

\input{6-implementation.tex}

\input{8-evaluation-v2.tex}

\input{9-related.tex}

\input{910-discussion}

\noindent\edited{\textbf{Acknowledgements--} { We thank our shepherd, anonymous reviewers, and all WCSNG lab members for their valuable suggestions and feedback.}}

\bibliographystyle{abbrv}
\bibliography{main}

\end{document}

%% file: 0-abstract.tex
\begin{abstract}
  Autonomous perception requires high-quality environment sensing in the form of 3D bounding boxes of dynamic objects. The primary sensors used in automotive systems are light-based cameras and LiDARs. However, they are known to fail in adverse weather conditions. Radars can potentially solve this problem as they are barely affected by adverse weather conditions. However, specular reflections of wireless signals cause poor performance of radar point clouds.  We introduce \name, a system that combines data from multiple spatially separated radars with an optimal separation to mitigate these problems. We introduce a novel concept of \textit{\pcname}, which uses the spatial diversity induced by multiple radars and solves the problem of noise and sparsity in radar point clouds. Furthermore, we present the design of \textit{\network}, a novel deep learning architecture, designed explicitly for radar's sparse data distribution, to enable accurate 3D bounding box estimation. The spatial techniques designed and proposed in this paper are fundamental to radars point cloud distribution and would benefit other radar sensing applications. Dataset available at project webpage: \url{https://wcsng.ucsd.edu/pointillism/}

\end{abstract}

\keywords{mmWaves, Radar Perception, Deep Learning, Adverse Weather, Autonomous Driving, Object Detection}

%% file: 1-intro-v8.tex
\vspace{-0.2cm}
\section{INTRODUCTION}

Autonomous vehicles require a high-quality geometric perception~\cite{van2018autonomous,darms2008classification,3dperception,3dperception2} of the scene in which they are navigating, even in adverse weather conditions. Most of the existing computer vision algorithms and data-driven techniques rely on high resolution, multi-channel LiDARs~\cite{perception,shi2018pointrcnn,wang2019frustum,Qi_2018} to construct accurate 3D bounding boxes for dynamic objects. LiDAR is a light-based sensor which measures the reflection of the light signal to perceive the geometry of the scene, creating dense 3D point clouds. 
However, LiDAR cannot penetrate through fog and dust, causing the sensor to fall prey to adverse weather conditions~\cite{dmv,bad_weather,forbes}. In contrast, radars are a robust sensing solution, which transmits millimeter waves (mmWaves) and remain less affected by adverse weather conditions~\cite{lidar_radar}. The wavelength of mmWaves allows them to easily pass through fog, dust, and other microscopic particles.

\begin{figure}[t]
\centering
\includegraphics[scale=0.4]{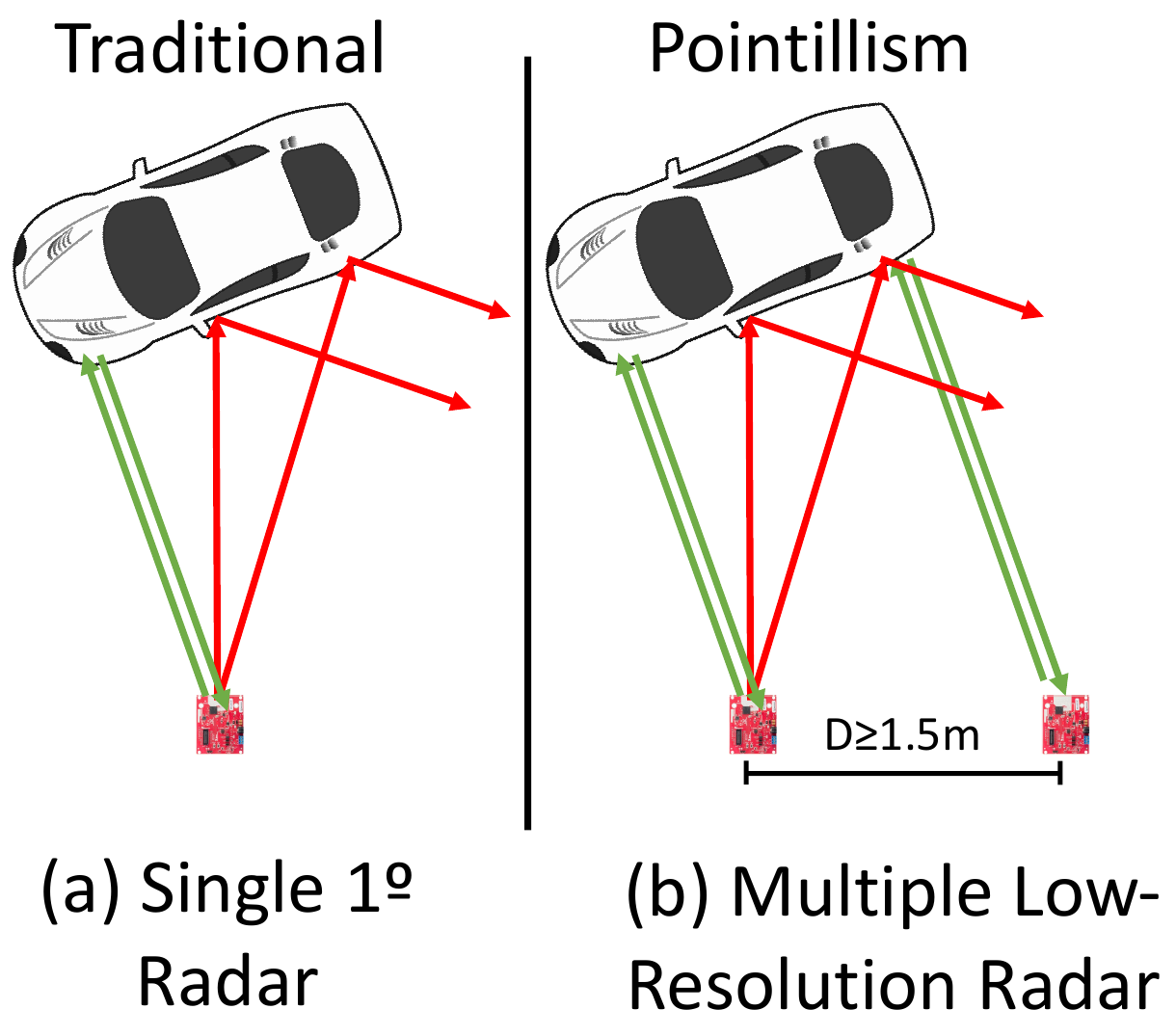}
\caption{ a) Single radar misses the reflective surface due to specular reflections (red). b) Multi-radar setup with optimal separation receives reflections back almost surely from the reflective surface (green).} 
\label{fig:specular}
\squeezeup
\vspace{-0.05cm}
\end{figure}

Although radars are all-weather reliant sensors, they need to provide LiDAR-like high-quality perception performance to enable adverse weather perception. However, a challenge with the radar is that it cannot create dense point clouds like LiDAR. The primary reason is that an automotive radar emits mmWave signals, which reflect specularly off of surfaces unlike light signals which scatter in every direction, allowing only a fraction of incident waves to travel back to the radar receiver, as shown in Figure~\ref{fig:specular}. Even a high angular-resolution automotive radar suffers from the curse of specularity and would only create a sparse point cloud with insufficient information for precise bounding box estimation. To further exacerbate the issue, radar data contains structured noise due to sensor leakage, background clutter, and multi-path effects. The noise pollutes the point clouds by causing unwanted points to appear in the scene point cloud. This leads to inaccuracies in identifying the number of objects in the scene by increasing false detections. 

In this work, we propose \name, a system that enables radars to overcome the challenges posed by specular reflections, sparsity and noise in the radar point clouds, to provide high-fidelity perception of the scene with 3D bounding boxes. \name consists of multiple low-resolution radars placed in a optimal fashion to maximize the spatial diversity and scene information. \name combines this spatial diversity with novel multi-radar fusion algorithms to tackle the problem of specular reflections, sparsity and noise in radar point clouds. Building upon the hardware and algorithms, \name also introduces a novel data-driven approach that enables the detection of multiple dynamic objects in the scene, with their accurate location, orientation and 3D dimensions. Furthermore, \name enables such perception even in inclement weather, thereby paving a way for radar to be the main-stream sensor for autonomous perception.

A natural question is how does \name overcomes the physics of wireless signals i.e., specularity. \name's key idea to overcome specular reflections is to use multiple radars placed at spatially separated locations overlooking the same scene and illuminate an object in the scene from different viewpoints. This, in turn, increases the probability of receiving a reflection back from multiple points/surfaces of the object, which single radar could have missed, as shown in Figure~\ref{fig:specular}. \name formulates the problem of optimal placement of multiple low-resolution radars as an optimization problem to achieve multiple reflection points from a vehicle at all orientation. The optimization reveals that the optimal radar placement of around 1.5 meter apart (typical width of a car) to achieve high-fidelity in the estimated pose of the surface.

A good placement ensures spatial diversity in the point clouds, but noise is still a major challenge. \name presents a novel multi-radar fusion algorithm that reduces the noise points to enable accurate detection of multiple dynamic objects. The insight is to use spatial diversity generated by multiple radars to reduce the noise and enhance points corresponding to actual dynamic objects (eliminate noise). A naive approach to leverage spatial diversity would be to translate the point clouds collected by each radar of the multiple-radar to a common frame of reference and combine them to densify the radar point cloud. However, this approach adds up the noise points as well, doesn't reduce the noise and misses out on the crucial information encoded in the spatial locations of radars. 

We make a key observation that across multiple viewpoints (radars), the noise points appears independent of each other in space and the points belonging to actual surface/object appear at nearby location consistently in most of the views (radars). To best leverage the observation, we create a space-time coherence based framework for combining of 3D point clouds from multiple radars. The output is a novel representation of \textit{\pcname} that have the information regarding the confidence of each point coming from an actual object as soft probability value, along with all the properties of a point cloud. 

With the knowledge of confidence estimates for the points, we can infer whether they belong to objects or noise. However, merely identifying all the relevant points out of noise is not sufficient for multi-object 3D bounding box estimation. Firstly, depending on the distance, orientation, and the exposed surface of an object, only a limited set of points could be captured by the radar. Secondly, in a scene with multiple objects, precise 3D bounding box estimation requires segmenting out the points belonging to each object. This results in massive uncertainty in the exact orientation and the location of the bounding box. 


A naive approach to solve for uncertainty could be to design hand-crafted features by taking into account the shape and size of the vehicles and all possible orientations. However, such an approach is not trivial because crafting the features that can incorporate all possible cases is very challenging. Our insight here is that a data-driven method could potentially solve this problem by building experience over time and learn the non-uniform distribution of radar point clouds.

We reap the advantages of data-driven learning for precise 3D bounding box estimation and propose a novel deep learning-based approach \textit{\network}, that leverages the sparsity of \pcname. \textit{\network} combines the two problems of point cloud segmentation and 3D bounding box location estimation in space, and performs a region of interest (RoI) based classification. However, picking RoIs uniformly throughout the 3D space is not computationally feasible. So, we define a unique set of anchor boxes that allow us to iterate over all the possible configurations of bounding boxes over sparse point clouds. Further, our experiments show that with this set of anchor boxes, we can exhaustively cover all the configurations while efficiently reducing the search space. Our solution is an end-to-end trainable network, which can easily be deployed for real-life testing.
 
It is well known that a data-driven approach needs a lot of data for training and validation. To the best of our knowledge, no publicly available dataset contains data from multiple low-resolution radars with overlapping field of views. Due to the lack of data, we built an automated data collection platform with multiple radars along with a LiDAR and a depth camera for ground truth labels of scene.

\name achieves a median error of less than \textbf{37cm} in localizing the center of an object bounding box, and a median error of less than \textbf{25cm} in estimating the dimensions of the bounding boxes. Moreover, \name achieves an overall mAP score~\cite{Geiger2012CVPR} of \textbf{0.67} with an IoU threshold of 0.5 and \textbf{0.94} with an IoU threshold of 0.2 for estimating 3D bounding box, which is comparable with the state-of-the-art bounding box estimation techniques~\cite{Qi_2018,shi2018pointrcnn,shi2019pv} using LiDARs. Furthermore, with our approach, the mAP values increase to \textbf{0.67} compared to \textbf{0.45} for a single radar system. This means that \name improves the performance by \textbf{48\%} with its multi-radar fusion compared to a single radar. \network can make inference at a frame rate of 50Hz which is well beyond the real-time requirements.

In summary, our contributions are as follows:
\vspace{-0.12in}
\begin{itemize}
    \item We propose \name, a new framework for radar perception that leverages spatial diversity induced by multiple radars and optimizes their separation, to counter the fundamental challenge of specular reflections in mmWave radars.
    \item We conceptualize the notion of \textit{\pcname} by utilizing space-time coherence on point clouds from multiple radars, to reduce the noise in radar point clouds, thereby increasing quality of signal.
    \item We provide the design of \textit{\network}, a novel deep learning framework, designed to leverage the non-uniform distribution of radar point clouds, and estimate precise 3D bounding boxes on \pcname.
    \item We build the first real-world dataset of labeled radar point clouds from multiple radars with overlapping field of view, that contains \datasize radar frames and corresponding LiDAR point clouds and RGB images in good/bad weather conditions. We believe that this dataset would enable innovation in a variety of sensor-fusion approaches.
\end{itemize}

%% file: 2-challenges.tex
 \section{Background and challenges}
In this section, we briefly describe the point cloud generation process used in the current automotive radars and outline the challenges involved in working with radar point clouds.  

\subsection{Radar Processing Pipeline}\label{sec:radar_processing}

FMCW (Frequency Modulated Continous Wave) technique has become a standard in the automotive radar market for point cloud generation and velocity estimation~\cite{TIfmcw,astyx}. Additionally, multiple transmit and receive antennas are used for the angle of arrival estimation using beamforming. We would refer the reader to~\cite{TIfmcw} for a detailed description of the entire point cloud generation. The resolution in range and angle estimation in FMCW depends on the bandwidth of operation and the number of available antennas, respectively. 

The current market trend is to achieve higher angular resolution by using several co-located antennas and improve bounding box estimation performance. However, radar point clouds are affected by more fundamental challenges that inhibit the bounding box estimation performance on radar data. In the next section, we describe the challenges with radar point clouds and discuss how they affect the bounding box estimation performance.

\begin{figure}[t]
\centering
\includegraphics[width=0.9\linewidth]{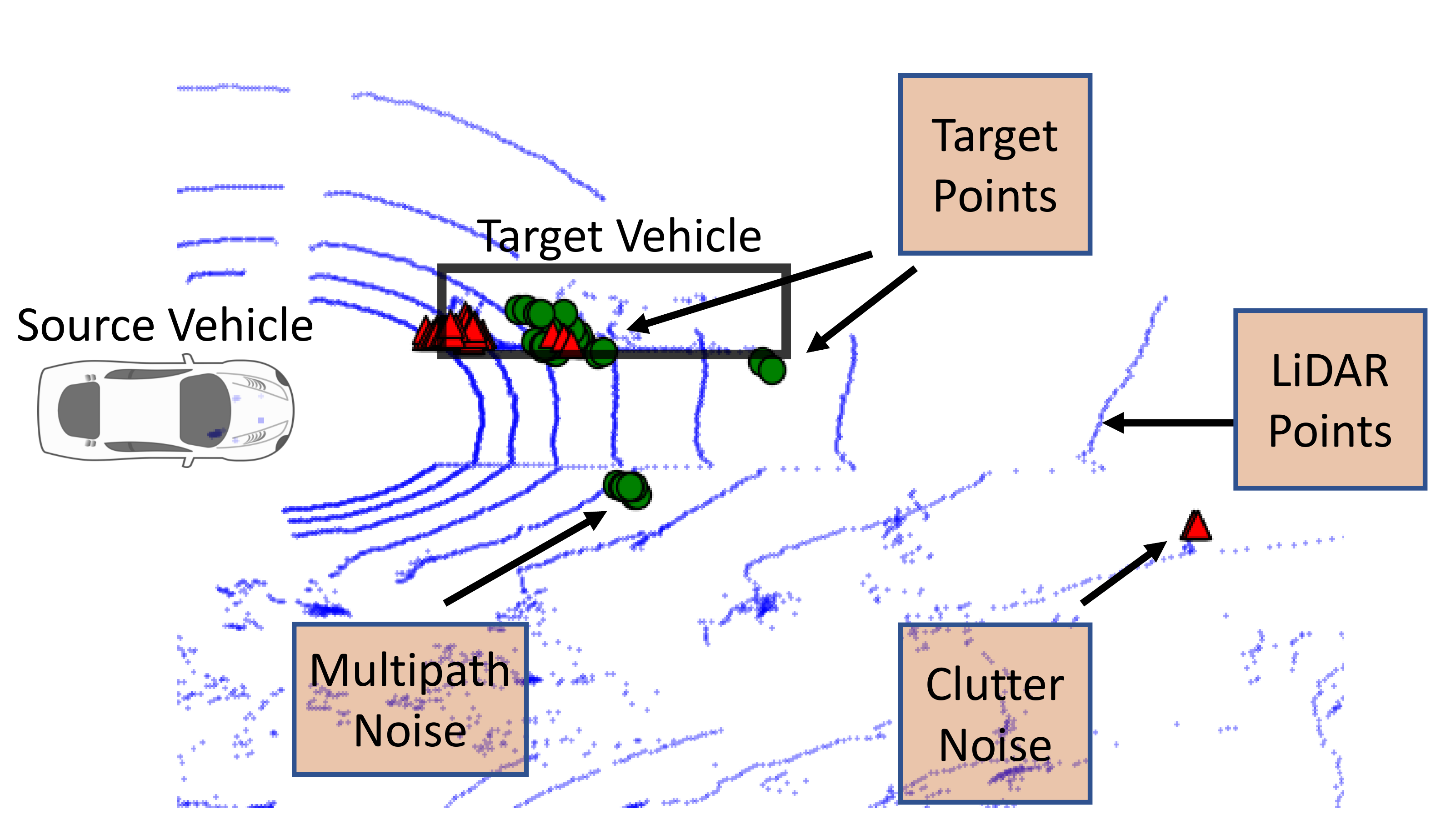}
\caption{Point-cloud of a scene with one target car using multiple radars (red triangles and green dots are two different radars, blue points are from LiDAR). Multiple such target vehicles could be present in the scene. Figure shows noisy radar points generated due to clutter and multipath effects, and their independence across two radars.}
\label{fig:noise_sparse}
\vspace{-0.7cm}
\end{figure}
\subsection{Challenges in radar point clouds}
Millimeter wave sensing has enabled high resolution radars for automotive sensing but it has its own perils. There are three main challenges faced in mmWave sensing: 


\noindent \textbf{(a) Specular reflections:}
For an incident electromagnetic wave on a surface, the size of its wavelength compared to the roughness of the object's surface determines the degree of scattering of the wave. mmWaves undergo a negligible scattering effect, resulting in a specular reflection (angle of incidence = angle of departure) from the surfaces. Consequently, for a small aperture radar, a lot of reflected signal does not make its way back to the sensor, causing blindness of the objects. This blindness is even independent of the resolution capabilities of the sensor.

\noindent \textbf{(b) Radar clutter and noise:}
Radar detections are commonly known to be polluted by signals from clutter, noise, and multipath effects. Radar clutter is defined as the unwanted echos from the ground or other objects like insects that can be confused with the objects under consideration. In a congested environment like cities, a signal emitted by a radar sensor could suffer multiple reflections before coming back to the sensor. The result is the formation of ghost objects, which are reflections of actual objects in some reflector formed because of multipath(Figure~\ref{fig:noise_sparse}).

\noindent \textbf{(c) Sparsity:}
Outdoor scene point clouds are inherently sparse due to the empty volume between the objects, which are at a substantial distance from each other. Additionally, due to different interaction properties of mmWaves with different objects (non-uniform interactions), this effect is compounded in the case of mmWave radars. The result is a sparse and non-uniform point cloud (Figure~\ref{fig:noise_sparse}).

%% file: 3-singleVsmultiple-v2.tex

\begin{figure}[t]
    \centering
    \includegraphics[width = 7cm]{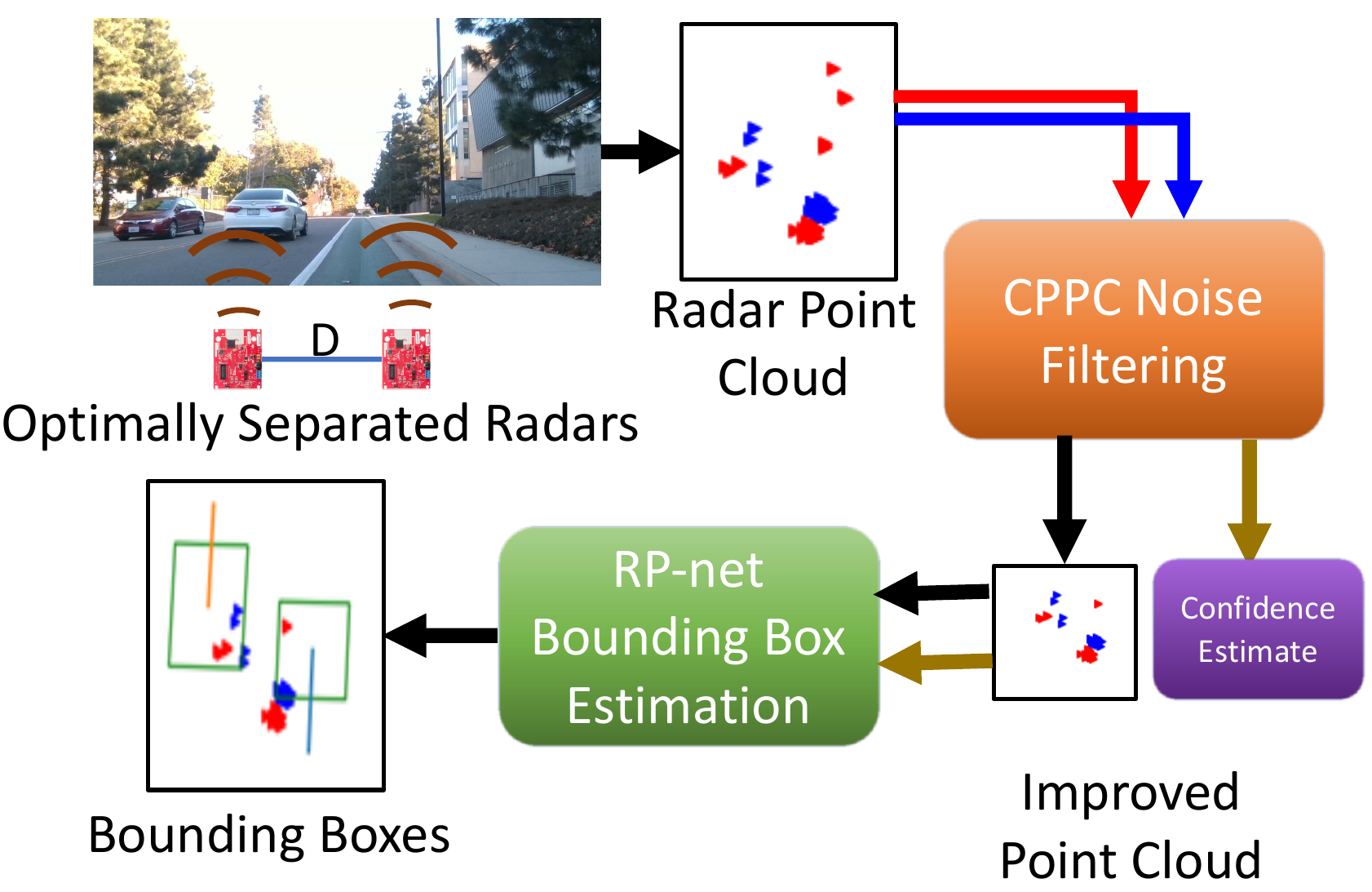}
    \caption{Overview of \name. PC: Point Cloud 
    }
    \label{fig:overview}
    \squeezeup
\end{figure}

 \name aims to tackle each of these challenges and provide accurate bounding boxes. Our approach towards achieving this goal is to couple multiple radar fusion with a novel noise filtering algorithm and estimate the bounding boxes. Figure \ref{fig:overview} shows an overview of our entire processing pipeline.

\section{Multi-Radar Perception}
\label{sec:Multi-radar}
Specular reflections of millimeter waves can cause direct blindness of object surfaces, which could lead to fatal accidents. To better scrutinize the effect of specularity, we need to understand the distribution of a radar point cloud. Point cloud generated by a radar depends mainly on two aspects: \textit{geometry} of the scene and \textit{resolution} of the radar (figure~\ref{fig:geometricVsSystem}b). Importantly, the adverse effect of specular reflections is a \textit{geometric} shortcoming and can not be tackled simply by increasing the resolution of the radar. 


\begin{figure}[t]
    \centering
    \includegraphics[width = 0.9\linewidth]{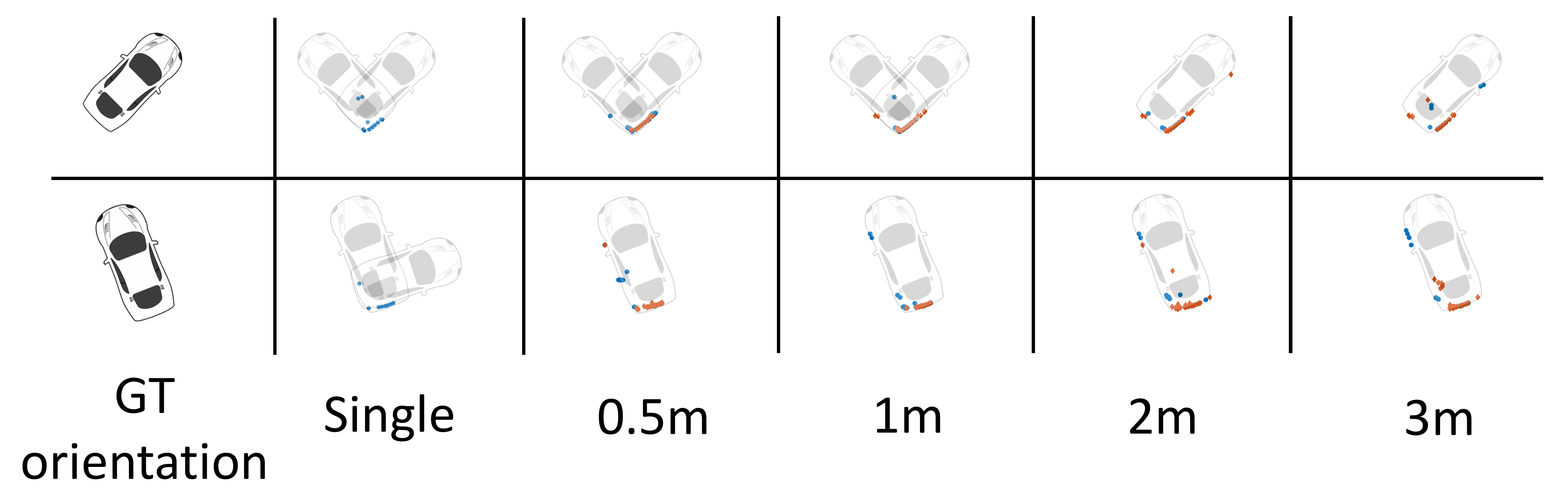}
    \caption{Wireless Insite simulations: Point clouds with single radar or smaller separation could lead to ambiguity in pose of the car which gets eliminated by increasing radar separation. Orange and blue points are from 2 different radars.
    }
    \label{fig:wi_sim}
    \vspace{-0.5cm}
\end{figure}

\subsection{Studying Radar point clouds}
To study the effect of scene geometry, we need to consider radars with a perfect resolution that can resolve any two reflections which are arbitrarily close to each other. We use a high-fidelity EM wave propagation tool Wireless InSite to model the EM wave interactions~\cite{WI}. Our experiments using a 1-degree radar (\cref{sec:evaluation}), and the results in past work~\cite{schuler2008extraction}, show that the radar reflection characteristics obtained from these simulations are similar to a real-world radar. We create simulations using a CAD model of a car along with its material properties. We create multiple simulations by placing the car at several locations in an area of 10m$\times$10m in front of the radar(s). At each location, simulations are created for multiple orientations(angles) of car, spanning 360 degrees in 36 discrete steps. For each placement and orientation, we consider two radar setups: \textit{Single radar} and \textit{Two radars with varying separation distance}. In simulations, the radars can resolve all the rays returning to the receiver (perfect resolution). With this, we can independently compare how the geometry of the scene affects performance. 

\subsection{Performance comparison of single and multiple radars}
\begin{figure}[t]
    \centering
    \includegraphics[width = 0.8\linewidth]{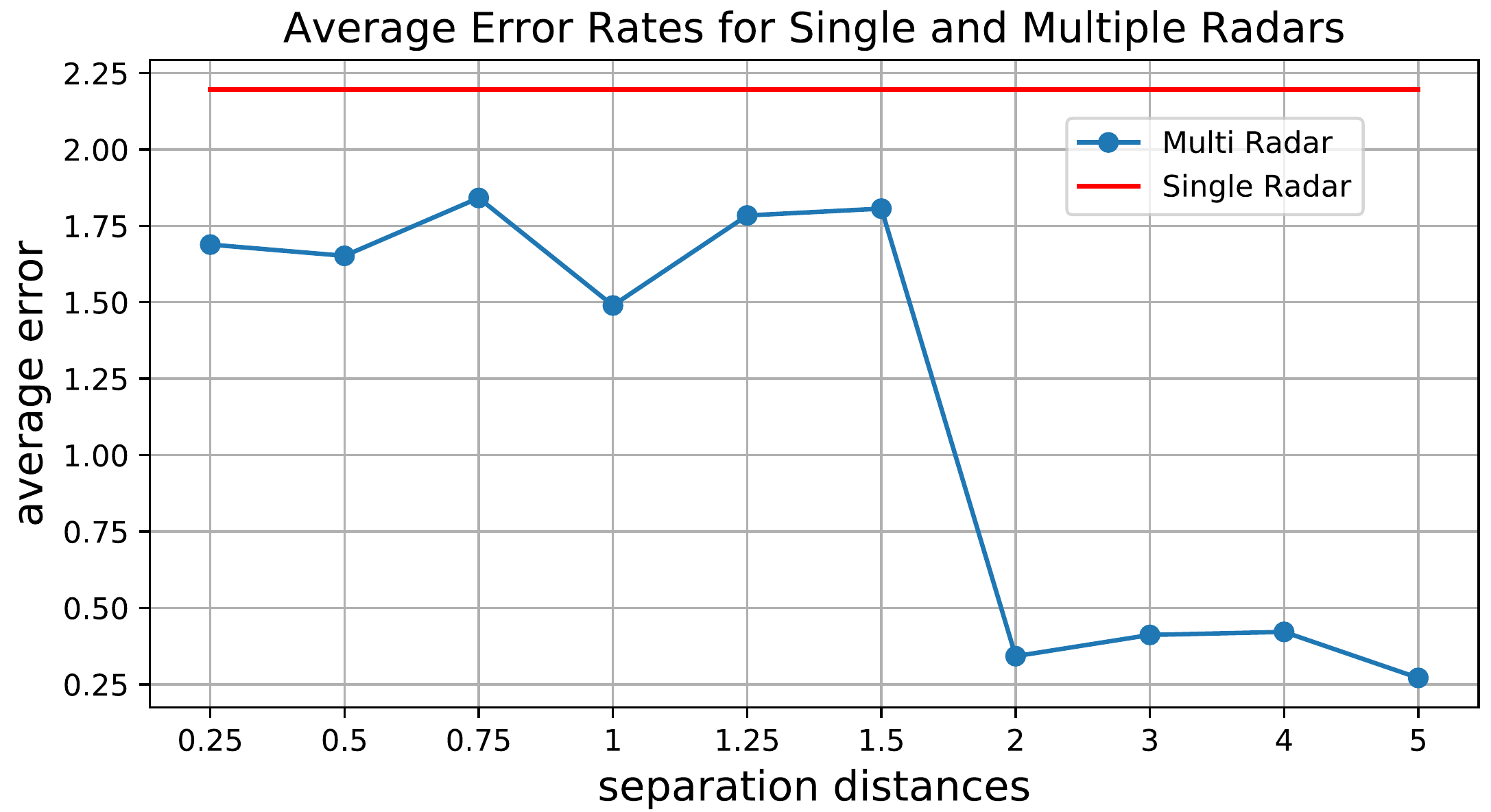}
    \caption{Comparison of average error in radian between single and multiple radars. For a separation greater than 1.5m, the performance improves.}
    \label{fig:simulations}
    \vspace{-0.5cm}
\end{figure}

For a rectangular bounding box of a car, the system that captures more surface points (less affected by specularity) would perform better in estimating the bounding box's orientation. To estimate the orientation angle of the box from point clouds, we use an MLP (multi-layer perceptron) regressor from \textit{scikit-learn}~\cite{mlpscikit-learn} trained on the data generated from the simulations. MLP regressor takes a simulated point cloud as input and outputs the orientation angle of the car. Figure~\ref{fig:simulations} shows the comparison of performance in terms of the mean error made in angle estimation. We make two important observations from these results. Firstly, the results show that the two radar system outperforms the single radar system. Secondly, a more interesting observation is that there is a sharp increase in performance between the separations 1.5m and 2m. It remains relatively constant before and after. The width of the car we used in simulations is $\approx$ 1.7m, which is the width of a standard car. Hence, the results show that the optimal distance between two radars for estimating a bounding box on a vehicle should be comparable to the vehicle's width. Figure ~\ref{fig:wi_sim} further shows how increasing the separation between radars would improve the point cloud generated.



\begin{figure}[t!]
    \centering
    \includegraphics[width = 0.9\linewidth]{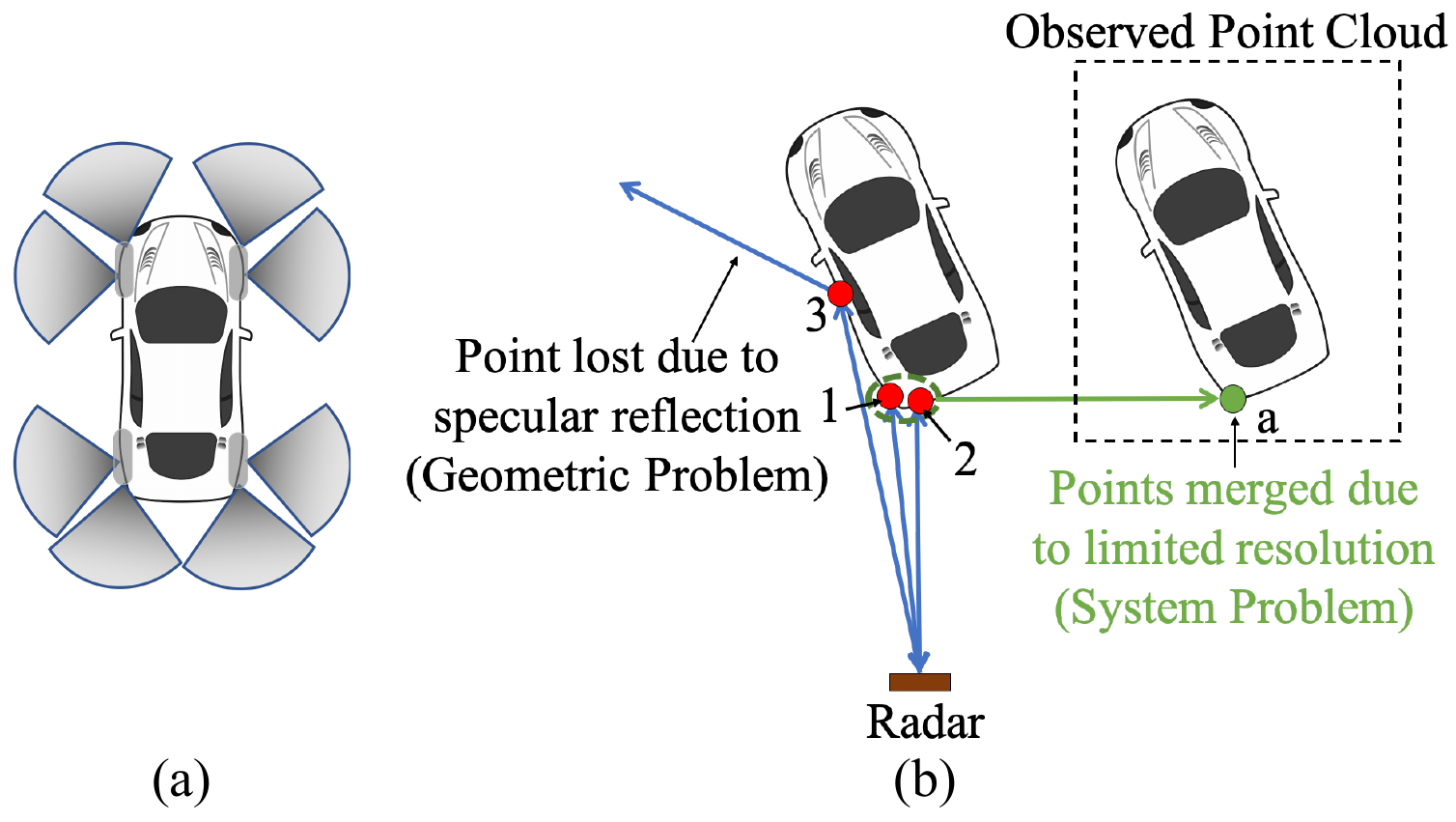}
    \caption{(a) Regions of reflection on a car. (b) Specularity vs Resolution:
     In the observed point cloud, point 3 is missing due to specular reflection while point \textit{a} is formed from points 1 \& 2 due to limited resolution.
    }
    \label{fig:geometricVsSystem}
    \vspace{-0.5cm}
\end{figure}

\textbf{What do multiple radars bring to the table?}
Based on our analysis following are the key benefits of using multiple radar system:

\begin{itemize}[leftmargin=*]
    \item \textbf{Larger Virtual Aperture:}The reflections arriving from a car originate from specific scattering points around the car (e.g., wheelhouses, corners)~\cite{buhren2006automotive,hammarstrand2012extended}. Each scattering point has a specific visibility region (Figure~\ref{fig:geometricVsSystem}a).  Multiple radars present at spatially separated locations create a larger virtual aperture, thereby capturing more of these reflecting centers. Occluded parts of an object from one viewpoint would appear in the other viewpoint, inherently increases the probability of capturing more meaningful points. This increase directly impacts the system's orientation estimation performance, as is evident from the results.
    \item \textbf{N times the number of points:} One obvious advantage of having  multiple sensors is the increase in the number of points, which would densify the sparse point cloud.
    \item \textbf{Rich Spatial Diversity:} A larger virtual aperture with multiple radars provides rich spatial diversity. In the next section, we would show how this spatial diversity could be used to curb noise by building confidence for the points in the point cloud.
\end{itemize}

%% file: 4-design-pcpotential-v2.tex
\vspace{-0.2cm}
\section{\pcname}
\label{sec:cppc}

Multiple radars work together to overcome the challenges posed by specular reflections of mmWaves by providing rich spatial diversity. However, noise (due to clutter, multi-path, and system noise) still presents a big challenge for object detection from radar point clouds. The noise points misguide object detection and introduce false positives. A single radar has fundamental limitations in removing noise. The features corresponding to the Cartesian coordinates, e.g. (x,y,z) and velocity, can provide a rich context for object detection. However, they do not provide information necessary to segregate the noise. 

In this section, we would introduce a novel representation of \textit{\pcname}, formed by the fusion of multiple radar point clouds. Our key observation is that points belonging to noise are independent across multiple radars placed at different spatial locations. In other words, points belonging to an actual object are more likely to be present in the point clouds of more than one radar. In contrast, the points belonging to random noise are specific to each radar. By leveraging this observation, we design a novel algorithm that filters noise from radar point clouds and creates low-noise \pcname. In the following subsection, we would describe how we can use the data from multiple radars to encode information regarding noise.



\begin{figure}
    \centering
    \includegraphics[scale=0.7]{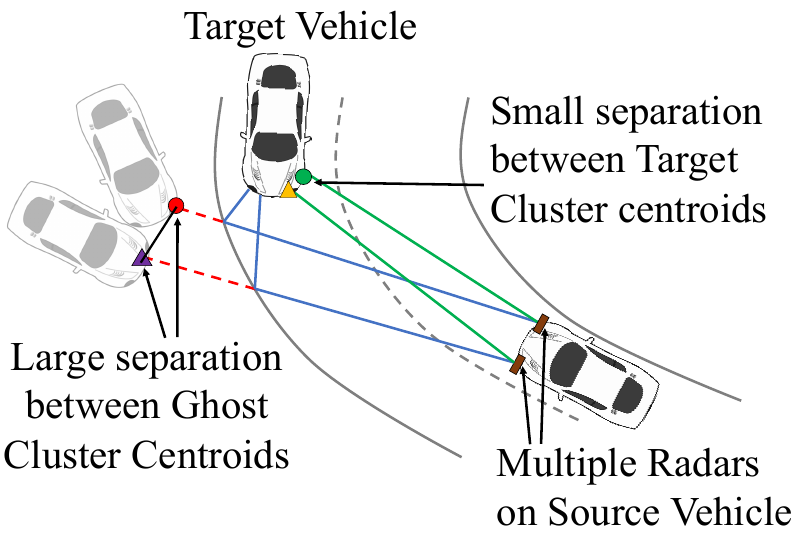}
    \caption{Formation of ghost clusters due to multipath. 
     The centroids in the ghost clusters are more separated in comparison to the centroid of target cluster and given low potential values in \pcname. 
    }
    \label{fig:noiseremoval}
    \vspace{-0.6cm}
\end{figure}
\vspace{-0.2cm}
\subsection{Space coherence with Radar point potential}
\label{subsec:space-time-coherence}
Noise harms the bounding box estimation as it creates false positives. Our key insight on tackling noise takes inspiration from signal processing techniques. In signal processing, we collect multiple noisy data stream and take their average to reduce the noise variance and improve the overall SNR (signal to noise ratio). The idea is that the signal present in each data stream adds up coherently. At the same time, the noise is random and will not add constructively. While the idea is intuitive, it is non-trivial to extend it to the point clouds. We cannot simply add point clouds from multiple radars in the hope to reduce noise because of the following reasons: (1) The 3D point cloud is sparse and incoherent in space, i.e., multiple radars may capture different points in 3D space for the same target object. (2) It is hard to build confidence for every point, whether it contributes to the object bounding box or corresponds to the noise point (generated by clutter or multi-path).





To apply the space coherence in the point cloud domain, we use the geometric information of point clouds. The above insight propagates to the fact that if a region of 3D space generates a response in multiple radars, it is likely to be generated from an object and not noise. To capture this effect, we need to measure the coherence between point clouds originating from multiple radars across 3D space. As shown in Figure~\ref{fig:geometricVsSystem}, radar points from an object are clustered around some scattering regions on a vehicle. By identifying these clusters, we can define our confidence of a point being generated from an object by looking at the same scattering region in multiple radar point clouds (space coherence). Our approach runs in two steps: (a) Clustering the point clouds and (b) Enforcing space coherence by defining cross-potentials, detailed as follows:


\noindent
\textbf{(a) Clustering the point clouds:}
Radar point clouds are present in the form of clusters of points originating from a scattering region on the object (Section \ref{sec:Multi-radar}). We use the standard DBSCAN~\cite{ester1996density} algorithm to find clusters in our point cloud. DBSCAN works on the notion of defining a neighborhood of points based on distance $\epsilon$ given as an input parameter. If a specific number of points (another input parameter) is present in that neighborhood, the point and its neighborhood are identified as a cluster. For each cluster $i$, the centroid $c_i$ of its points is used as the cluster's representative point.  For the multi-radar case, we use $c_{i}^{j}$ to represent the centroid of cluster $i$ in the point cloud of $j$th-radar, which is represented by $\Gamma_{j}$. In this way, we create multiple clusters individually for each radar point cloud.

\noindent
\textbf{(b) Cross-potential:} Next, we discuss how we find the correspondence between clusters across multiple radars. We define \textit{cross-potential} between two clusters from two different radars as a confidence metric if the two clusters belong to the same object. Intuitively, the cross-potential between two clusters is inversely proportional to the distance between the two clusters' centroid. We denote the cross-potential as $P(c_{i}^{j}|\Gamma_{k})$ for the $i$th cluster in radar $j$ ($c_i^j$ denotes its centroid) with the $k$th radar   for $k{\neq}j$, $k\in\{1,2,\ldots,N\}$ in an $N$-radar system. Mathematically, we define $P(c_{i}^{j}|\Gamma_{k})$ as




\begin{equation}
    {P}(c_{i}^{j}|\Gamma_{k}) = \frac{1}{1+\left[\frac{r_{i}^{jk}}{2}\right]^{2}}    
\end{equation}
where $r_{i}^{jk}$ is the distance between centroid $c_{i}^{j}$ from the respective nearest cluster centroid in $k$th-radar's point cloud $\Gamma_k$. 
\begin{figure}[t]
\centering
\includegraphics[width=1\linewidth]{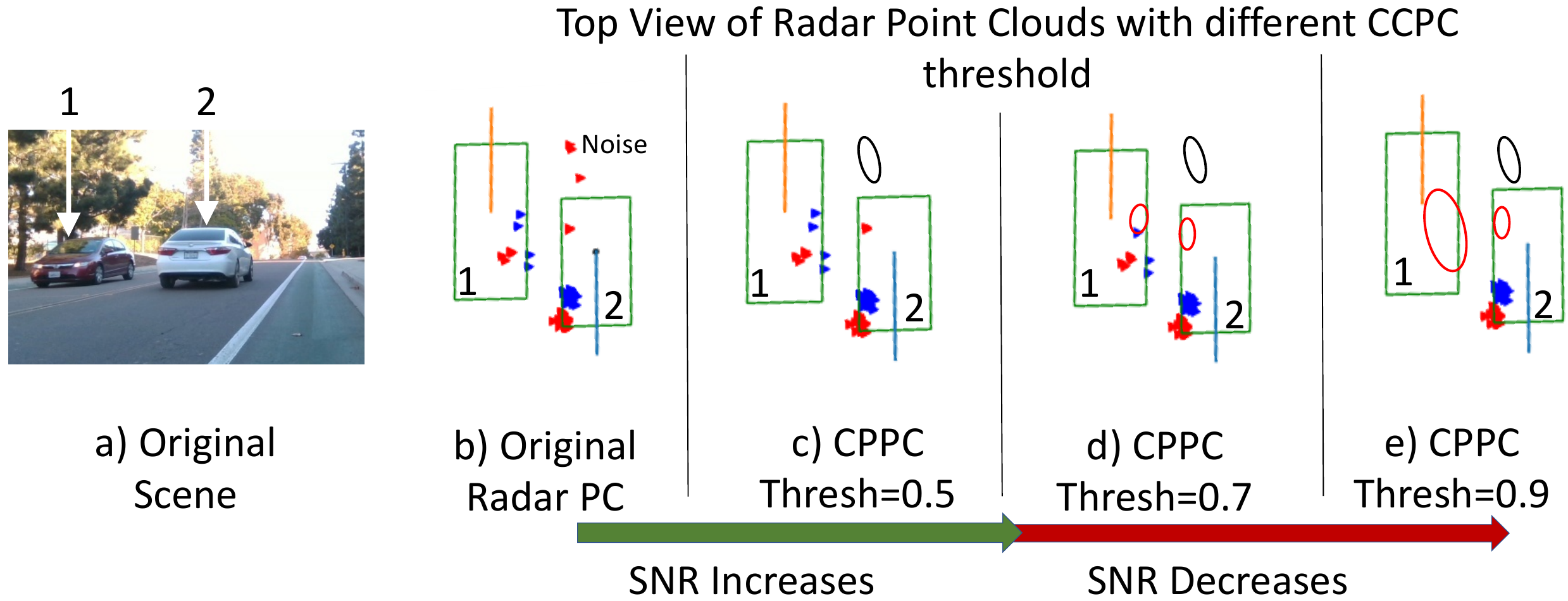}
\caption{Scene with multiple cars and effect of changing potential thresholds. Black (or red) ellipses show regions from where noise (or actual) points get filtered. At a very high threshold (0.9), all points from one car get filtered out, causing miss-detection. The blue and orange line shows the direction of car.}
\label{fig:cppc_snr_examples}
\vspace{-0.2 in}
\end{figure}

The reason behind this specific choice of potential is motivated by the dimensions of a car. A typical car has a width of around 2m. Therefore, we choose a function that gives a high potential to all the points within the 2m neighborhood of a point, i.e., \textsc{P}>0.5 if $r_{i}^{jk}$<2m. Note that our choice of potential function also preserves a point lying on a vehicle, which is present in only one radar, as long as it is in the 2m neighborhood of another high potential point (possibly on the farther corner along the width). With this, we preserve the extra points added due to the spatial diversity of the multiple radar system (Section \ref{sec:Multi-radar}).
Using this potential function, we can quantify the \textit{space coherence} of signals coming from multiple radars. We combine all the points from multiple radar point clouds and add confidence information to them. Each point gets the same potential as its respective cluster centroid. Further, we filter all the points below a certain potential-threshold (discussed in detail later) to create \textit{\pcname}.  (We would refer to this as CPPC algorithm for brevity). Figure~\ref{fig:noiseremoval} shows how CPPC algorithm gives a low potential to noise or multipath points.

\begin{figure}[t]
\centering
\includegraphics[width=0.8\linewidth]{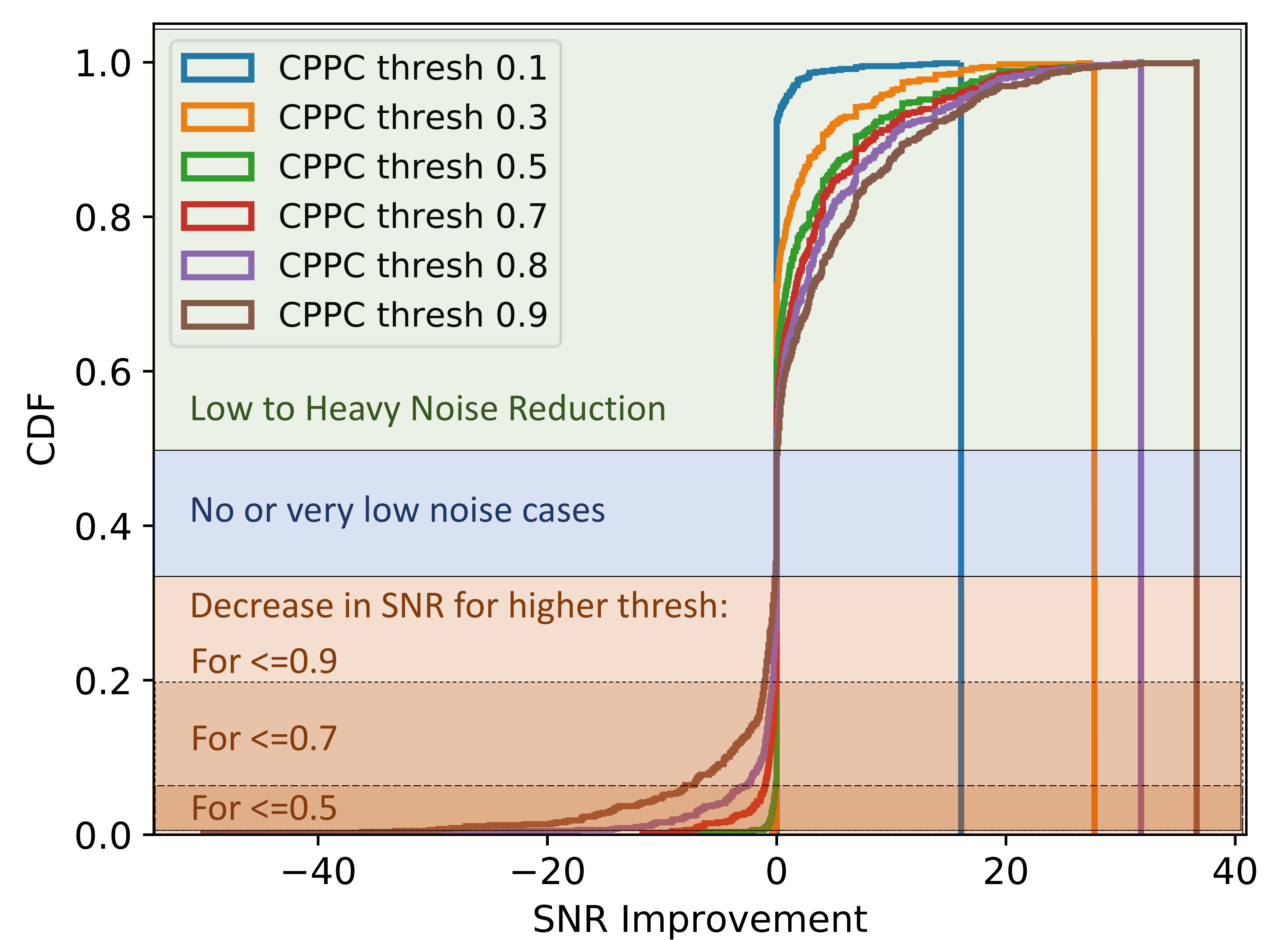}
\caption{SNR Improvement: Comparison of SNR improvement over base case (without CPPC) for multiple potential-thresholds. Higher threshold rejects more noise (green) with diminishing returns. Blue region shows the cases with very low noise in base case, hence there is no improvement. Red region shows the cases where SNR decreases severely for higher thresholds.}
\label{fig:cppc_snr}
\vspace{-0.2 in}
\end{figure}

To study the effect of different threshold values on CPPC, we define SNR (signal-to-noise ratio) of a point cloud as the ratio of the total number of actual points against the noise points. Noise points are defined as the points lying outside the bounding box of the vehicle. Figure \ref{fig:cppc_snr} shows the CDF plot of SNR improvement by using CPPC over the case where no CPPC transformation is applied (points from all the radars are combined without any filtering) for multiple potential thresholds. The plot shows that improvement in SNR increases with a higher potential threshold (green region) but with diminishing returns for large threshold values. Certain cases with low or no noise do not show improvement (blue region). However, a large threshold also decreases the SNR to a large extent in some cases, as it regards some actual points as noise and removes them, thereby reducing the SNR (Red region). The SNR decrease is severe for higher thresholds (>0.5) and could lead to missed-detections (Figure \ref{fig:cppc_snr_examples}). Hence, we choose 0.5 as our operating point to achieve a good trade-off between SNR improvement and missed-detections. 

\vspace{-0.2cm}
\subsection{Time coherence with multiple frames}
Space coherence can help to determine the noise out of the actual signal and boost detection performance. However, the task of estimating a bounding box is still non-trivial due to the sparsity of point clouds. Specifically, estimating the heading direction (also called pose or yaw angle of the bounding box) of a vehicle from a sparse point cloud within a single frame is challenging. In many cases, the points in a single frame might not contain enough information to reveal the heading direction.

To solve this problem, we make an important observation that along with the space coherence across the radars, the points also follow a \textit{time coherence} across multiple frames from radar. For a point originating from a rigid body, the linear motion would be the same as the car's motion across multiple time frames. We track the movement of points in consecutive frames and use it to estimate heading direction. Tracking is performed on self-motion compensated frames to remove the effect of the source vehicle movement. Kalman filter-based corrections are employed to tackle sensor uncertainties and noise. We track the points with the highest cross-potentials for each object. Using the time coherence between points, we can get a prior estimate of vehicles' heading directions in the scene.
\vspace{-0.2cm}

%% file: 5-design-LLPnet.tex
\section{Multi-Object 3D bounding boxes}

Using multiple radar fusion to create \textit{\pcname}, we can tackle noise and potentially eliminate in-accuracy in the detection of the number of dynamic objects. However, estimating the number of dynamic objects and their bounding boxes is still not a trivial task. To understand this, we need to look into the mapping function of radar. Given a scene geometry of multiple vehicles and environment, a radar maps it to point cloud information as
\begin{equation}
x,y,z = \mathcal{F}(\text{scene geometry})
\end{equation}
Our aim is to inverse this mapping and estimate the scene geometry in terms of object bounding boxes. 
\begin{equation} \label{eq:detector_eqn_cppc}
    (p_N,\psi_N) = \mathcal{D}(\Gamma_{CPPC})
\end{equation}

where $N$ is the unknown number of objects present in the scene; $\Gamma_{CPPC}$ is \pcname where each point is denoted by its cartesian coordinates, velocity, intensity and CPPC confidence; $p_N$ represents the confidence of detection for $N$th objects's bounding box in a scene and $\psi_N: \{c_x,c_y,c_z,w,h,l,\theta\}$ denote the tuple of bounding box parameters which are center coordinates, dimensions and yaw angle (angle with respect to $z$-axis) respectively. $\mathcal{D}$ is the multiple 3D bounding box estimation system. Estimating 3D bounding boxes for objects from a radar point cloud has the following challenges:



\begin{figure*}[ht]
\centering
\includegraphics[width=\linewidth]{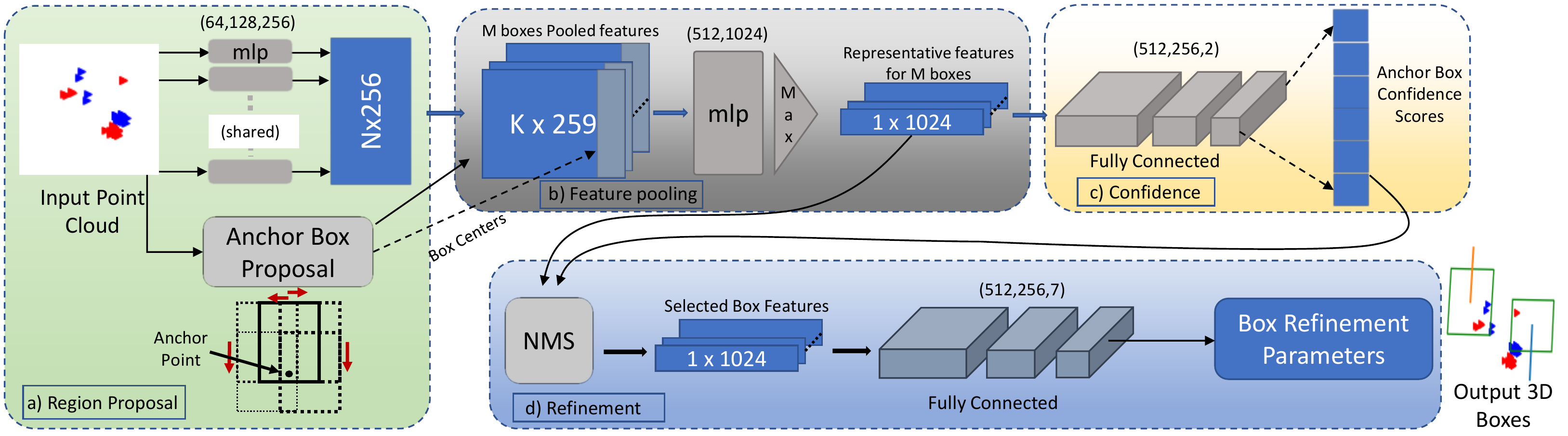}
\caption{\textbf{\network architecture:} a) Our network takes \textit{CPPC} as input with N points and 6 channels, extracts features using a pointnet layer, and generates anchor boxes for each point. b) A Feature pooling layer pools point features from K points lying inside each box, which are passed through pointnet and max pool layers to generate representative features for anchors. c) Confidence values are predicted for each anchor. d) High confidence anchors are taken to the box refinement stage. The output is a set of 3D bounding boxes for each object in the scene (top view box is shown).}
\label{fig:network}
\squeezeup
\end{figure*}

\begin{itemize}[leftmargin=*]
    \setlength \itemsep{-0.02in}
    \item \textbf{Segmentation:} The radar point cloud of a scene is sparsely distributed where any subset of points could belong to a single object. Also, the number of objects and their locations are not known a-priori. Bounding box estimation requires proper segmentation of points belonging to each object in the scene. This is a complex mapping problem where the number of targets is not known.
    \item \textbf{Uncertainty in bounding box parameters:} Radar can only see the part of an object which is exposed to the sensor. Therefore, the point cloud of an object may not contain crucial information regarding all dimensions, orientation, and center-location of the bounding box. As a result, there is uncertainty in bounding box parameters.
\end{itemize}


Estimating $\mathcal{D}$ with an exponentially large search space is a complex problem. A naive solution like fitting a fixed-sized bounding box based on point clusters will not work due to a lack of segmentation information. Proper segmentation of points into different objects and accurate 3D bounding box parameter estimation would require complex hand-crafted feature representations of the point cloud. However, recent advances in deep neural networks~\cite{qi2017pointnet,shi2018pointrcnn} have shown that we could overcome these challenges by learning complex feature representations from data.
In other words, a deep network can build experience regarding the relation between point clouds and exact object placements and use that experience to estimate accurate 3D bounding boxes.

\vspace{-0.2cm}
\subsection{\network Overview}
To overcome the challenges mentioned above, we present a novel deep learning architecture called \textit{\network}. Our network is specifically designed to handle the sparsity in radar point clouds and output accurate 3D bounding boxes. Directly giving the raw point cloud as an input can not solve the problem due to the unknown number and locations of the objects in the scene. Usually, an object detector uses a fixed set of initial region proposals to solve this problem. Our network proposes a novel way of generating these region proposals (anchor boxes), based on the radar response due to vehicle geometry and space-time coherence. Unlike LiDAR, where many points originate from ground and other static objects like buildings, radar data is sparse and contains mostly the points from dynamic and metallic objects like cars after CPPC noise suppression due to strong EM reflective properties of metals. Specifically, the sparsity in radar data and the fact that all the points originate from the vehicles' surface allow us to define point-based region proposals. These fixed size anchor boxes are used as initial estimates of 3D bounding boxes. The size of these anchor boxes is determined by the average size of vehicles in the training dataset. Now, instead of the entire scene point cloud, the network works separately on each of these anchor boxes. For each of these anchor boxes, the task is reduced to generate a confidence number $p$ of whether the points inside that anchor box belongs to an object.



Confidence scores are generated for all the anchor boxes. A set of high confidence boxes is chosen. These anchor boxes are passed through a refinement stage that solves the uncertainty issue in bounding box parameters. In this stage, the anchor boxes are refined to provide accurate 3D bounding boxes of the objects present in the scene (parameter $\psi$). 



\vspace{-0.25in} 
\subsection{Network Architecture Design}

The job of \network is that given a scene point cloud, output the set of 3D bounding boxes of the objects present in the scene. It takes CPPC as input, which has 6 channels corresponding to x,y,z coordinates, velocity, peak intensities, and cross-potential values for each point. \network comprises of following blocks:


\noindent \textbf{(a) Handling Multiple Objects.}For a given scene with multiple vehicles, region proposals are defined by placing multiple anchor boxes in the scene. As mentioned above,  we propose a novel point-based region proposal (anchor boxes) generation scheme based on the radar response due to vehicle geometry and space-time coherence. Given a point, we use five different placements of anchor box around the point (we call this point as \textit{anchor point} for those anchor boxes. Refer Figure~\ref{fig:network}).  We use the pose values derived for each anchor point using space-time coherence for the orientation angle, as explained in Section 4.

\noindent \textbf{(b) Segmentation by Feature Extraction and pooling.}
Our objective is to perform classification and 3D bounding box parameter regression by learning meaningful feature representations from the point cloud data. \network extracts these features in two stages, i.e., before and after generating \textit{anchor boxes}. We first use a pointnet~\cite{qi2017pointnet} encoder of shared MLP to extract features from the entire point cloud. We would refer the reader to~\cite{qi2017pointnet} for an in-depth understanding of pointnets. 

In the second stage, the anchor boxes are determined for each point. An RoI (Region of Interest) feature pooling block of the network pools the features from all the points lying inside an anchor box. These features are passed through another pointnet layer and then max pooled into a single representative feature for every anchor box defined per scene. 



\noindent \textbf{(c) Bounding Box confidence prediction.}
 The entire set of representative features of anchor boxes, obtained after the previous block, is passed through a classification network consisting of fully connected layers. The fully connected layers learn a mapping from anchor boxes' representative features to the confidence value for each box. 
 
 Performing classification on RoI based max-pooled features always ensures that the contextual information from all neighborhood points of the anchor point, lying inside the anchor box is accounted, leading to better classification results. The problem of segmentation is solved by performing classification directly on the anchor boxes. The network will learn to choose the corresponding anchor box with high confidence, which contains all the points belonging to an object.


\noindent \textbf{(d) Refinement of Box Parameters.}
In the previous step, the anchor boxes were rough estimates of the dimensions, center, and orientation of final 3D bounding boxes as we used fixed-size anchor boxes. We still need further refinement of these parameters to get accurate bounding boxes. Note that this step is quite essential to estimate the accurate dimensions and location of the boxes. After the classification step, we obtain the confidence scores for all the anchor boxes. Since we obtained anchor boxes for each point, there could be many overlapping high confidence boxes belonging to the same object. Non-maximal suppression(NMS) sampling is performed on this set using the confidence values. NMS sampling removes boxes which have a high overlap with another high confidence box of the same object. The representative features from the remaining anchor boxes are passed through three fully connected layers to output a tuple $[h',w',l',x',y',z',\theta']$ corresponding to refinements of length, breadth, height, center coordinates, and orientation angle respectively. These refinements are added to the anchor box parameters to generate the final 3D bounding box prediction.

\noindent \textbf{(e) Loss functions.}
The anchor box classification in the first stage of the network is a binary classification problem that uses a cross-entropy loss, given by 
\begin{equation}
    \mathcal{L}_{RPN} = \sum_{i=1}^{N}-(y_i\log(p_i)+(1-y_i)\log(1-p_i))
\end{equation}
where $y_i=[0,1]$ is the ground truth and $p_i$ is the predicted confidence value. Refinement of the bounding boxes is a regression problem and we use Smooth-L\textsubscript{1} loss for this purpose. The loss is given by:
\begin{equation}
  \mathcal{L}_{refinement}(r,r')=\begin{cases}
    \frac{1}{2}(r-r')^2, & \text{for $|r-r'|<1$}.\\
    \delta|r-r'|-\frac{1}{2}, & \text{otherwise}.
  \end{cases}
\end{equation}
where $r$ and $r'$ are ground truth and regressed refinement values respectively for each parameter $[h',w',l',x',y',z',\theta']$.

%% file: 6-implementation.tex
\section{Experimental Setup and Dataset } 

\begin{table}[h!]
\begin{tabular}{c|c||c|c}
\hline
\textbf{Parameter} & \textbf{Value} & \textbf{Parameter}  & \textbf{Value} \\ \hline \hline
Start Frequency    & 77 GHz         & Frame rate          & 30 fps         \\
Bandwidth          & 2240 MHz       & Range Res.    & 0.067m         \\
ADC rate           & 7500 ksps      & Velocity Res. & 2.59 m/s       \\
Chirp Duration     & 40 $\mu$s          & Max velocity        & 20.74 m/s      \\ \hline
\end{tabular}
\captionsetup{width=0.99\columnwidth}
\caption{Values of the radar parameters used. (Res.=resolution)}
\label{table:parameters} 
\vspace{-1cm}
\end{table}

\begin{figure}[t]
\centering
\includegraphics[scale=0.5]{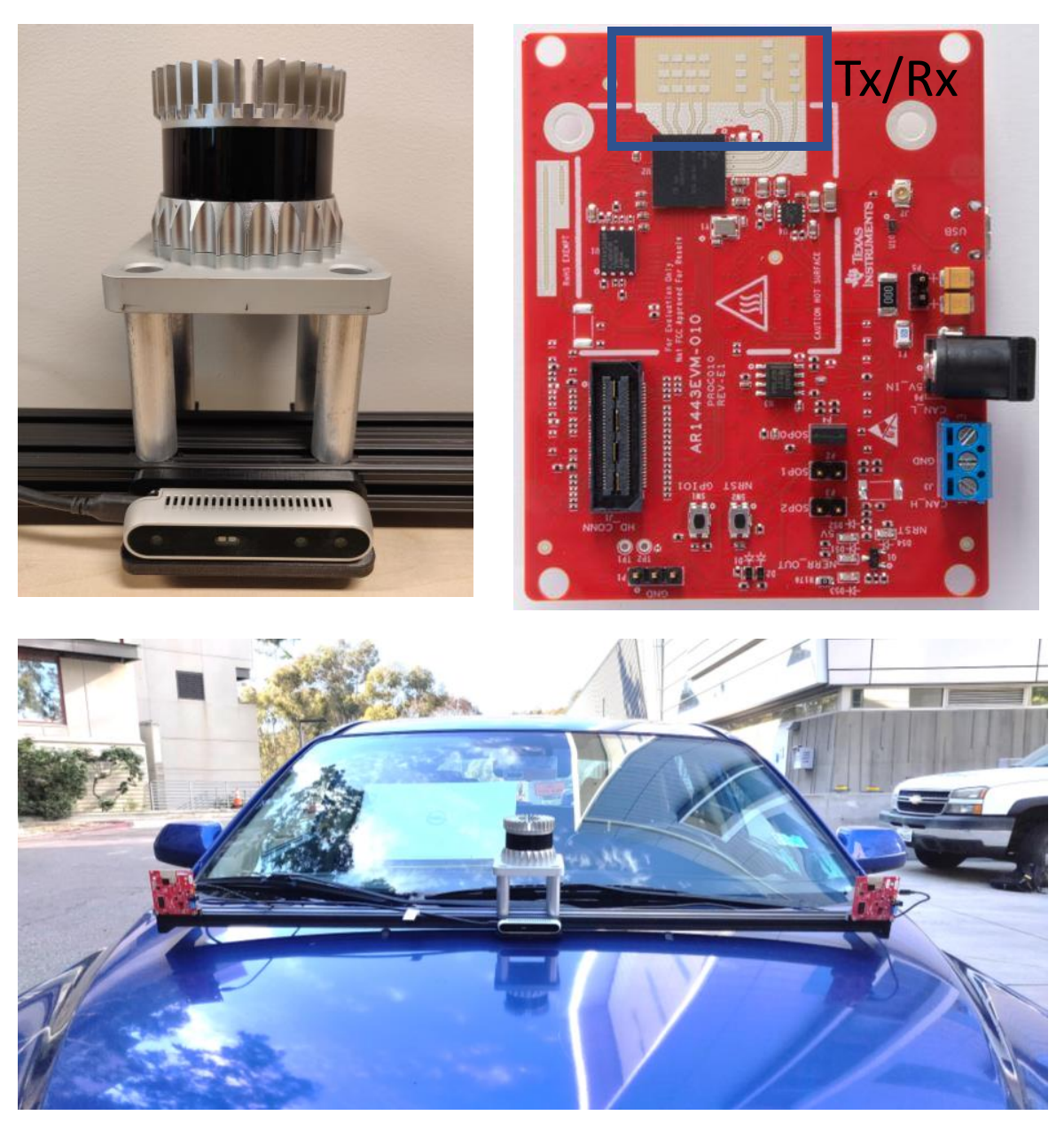}
\caption{\textbf{\name data collection platform}. (Top Left) 16-channel Ouster LiDAR, RealSense Depth Camera for ground truth labels. (Top Right) our radars. (Bottom) Our System: Two radars, LiDAR, Depth camera combined synchronously to common clock.}
\label{fig:hardware}
\vspace{-0.45cm}
\end{figure}

The use of radars for perception in autonomous systems is quite recent. No publicly available dataset collects data from multiple radar sensors with overlapping fields of view in 3D. We collect our own dataset for training and testing \name. We collected data from three sensors: a 16 channel Ouster LiDAR \cite{ouster} placed and an Intel RealSense D415 \cite{realsense} Camera at the center of a rail
, and 2 TI IWR1443BOOST \cite{TIfmcw} radars placed at the end-points of the rail at a distance of 1.5m, all placed on a car. Figure~\ref{fig:hardware} shows our data collection platform. We use LiDAR for ground truth annotations and camera for visualization only. Data is labeled using an online tool called \textit{scalabel}~\cite{scalabel}. All the sensors are controlled using a central controller running ROS (Robotic Operating System), running on a laptop placed in the car. Each data frame is timestamped for synchronizing the sensors. We perform controlled experimentation based sensor calibration. The data from all the sensors are brought to the same coordinate system by deriving extrinsic matrices for each sensor. Our hardware design is modular so that sensors can be swapped, and only data extraction for that sensor needs to be updated.

\begin{figure}[t]
\centering
\includegraphics[width=\linewidth,]{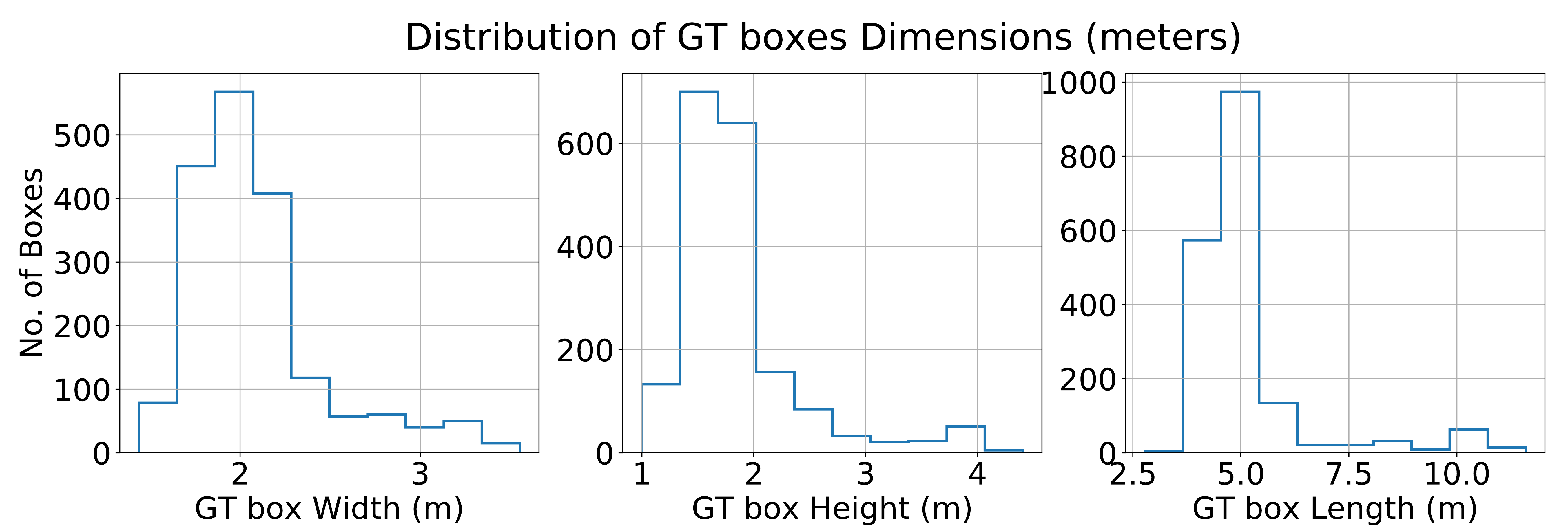}
\caption{Dataset Distribution: Histogram plots of the dimensions of the vehicles present in the dataset. It includes small golf carts to large buses.}
\label{fig:box_dist}
\squeezeup
\end{figure}

Each radar has a 3Tx and 4Rx MIMO antenna array. We use the FMCW processing chain provided by TI for generating point clouds characterized by x,y,z coordinates, doppler, and the intensity (peak values) of points~\cite{TIfmcw}. Table~\ref{table:parameters} summarizes the operating parameters of our radars. Figure \ref{fig:box_dist} shows the distribution of the different sizes of objects/cars present in our dataset. The dataset includes small golf carts to large buses with a maximum number of 4 objects in the scene. In total, we collect \datasize radar frames for five real-life traffic and driving scenes. Data was collected in different operating conditions, including day, night, and adverse weather (fog) conditions (smoke machine experiments). It is the first of its kind radar dataset with data from multiple radars, LiDARs, and RGB cameras that can enable various sensor-fusion-based approaches. We will release the dataset for public use. We plan to keep expanding this dataset with more samples over time.

\vspace{-0.2cm}
\section{Implementation}
In this section we would describe the implementation details of \name.

\noindent \textbf{Deep learning parameters:}
For implementing our network, we use the PyTorch framework. The number of output channels is mentioned above each layer in Figure~\ref{fig:network}. We use Adam optimizer with a learning rate of 0.0002 and momentum 0.9. A batch-normalization layer follows each layer of the network. We use (2m, 2m, 5m) as (width, height, length) to initialize the anchor boxes' dimensions. 70 random points are sampled from each \pcname for giving input to the network. The network takes 6 channel inputs for xyz-coordinates, velocity, peak intensity values, and confidence estimates. We sample 32 points at random from all the points lying inside an anchor box during the feature pooling stage. If the number of points is less than 32, then the same points are repeated to maintain consistency throughout network operation.

\noindent \textbf{End-to-end training and deployment:} For training the classification network, ground truth is generated based on the IoU (Intersection over Union) values of the anchor boxes against the ground truth bounding boxes. We use the top view of the anchor boxes to calculate 2D IoUs. We choose choose top 100 anchor boxes based on IoU match for classification. Boxes with 2D IoU overlap of greater than 0.2 are considered as positive examples for classification. For the refinement network, NMS is performed on the boxes to remove boxes with more than 0.5 2D IoU overlap using predicted confidence scores. The training happens simultaneously for both region proposal and the refinement network. However, at the start of the training, the classification head can not provide correct confidence scores; hence, for the initial 30 epochs, we use 2D IoU scores with ground-truth boxes as confidence values for NMS. Note that the network is already trained during the testing phase, and we do not need these IoU values.

\vspace{-0.3cm}

%% file: 8-evaluation-v2.tex
\section{Evaluation}
\label{sec:evaluation}

\begin{figure}[t]
\centering
\includegraphics[width=\linewidth]{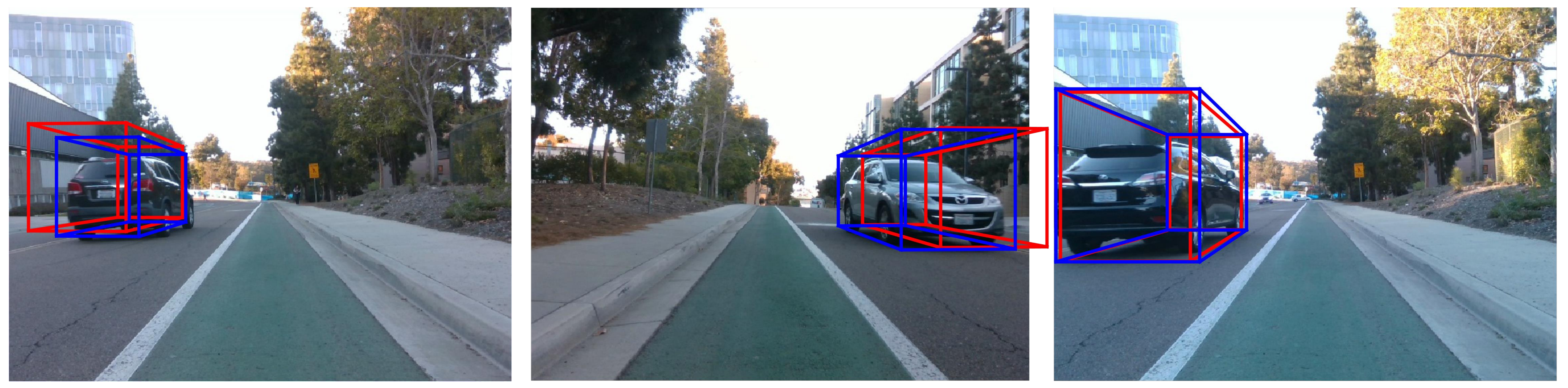}
\caption{Predicted 3D bounding boxes using only radar (red) along with ground truth boxes (blue). The 3D IoUs of predictions are 0.52, 0.43 \& 0.83 respectively. Radars can be used as a standalone sensor for object detection despite sparsity and low resolution.}
\label{fig:qualitative}
\vspace{-0.3in}
\end{figure}

\begin{figure*}[t]
\begin{minipage}{0.3\linewidth}
 \includegraphics[width=\linewidth]{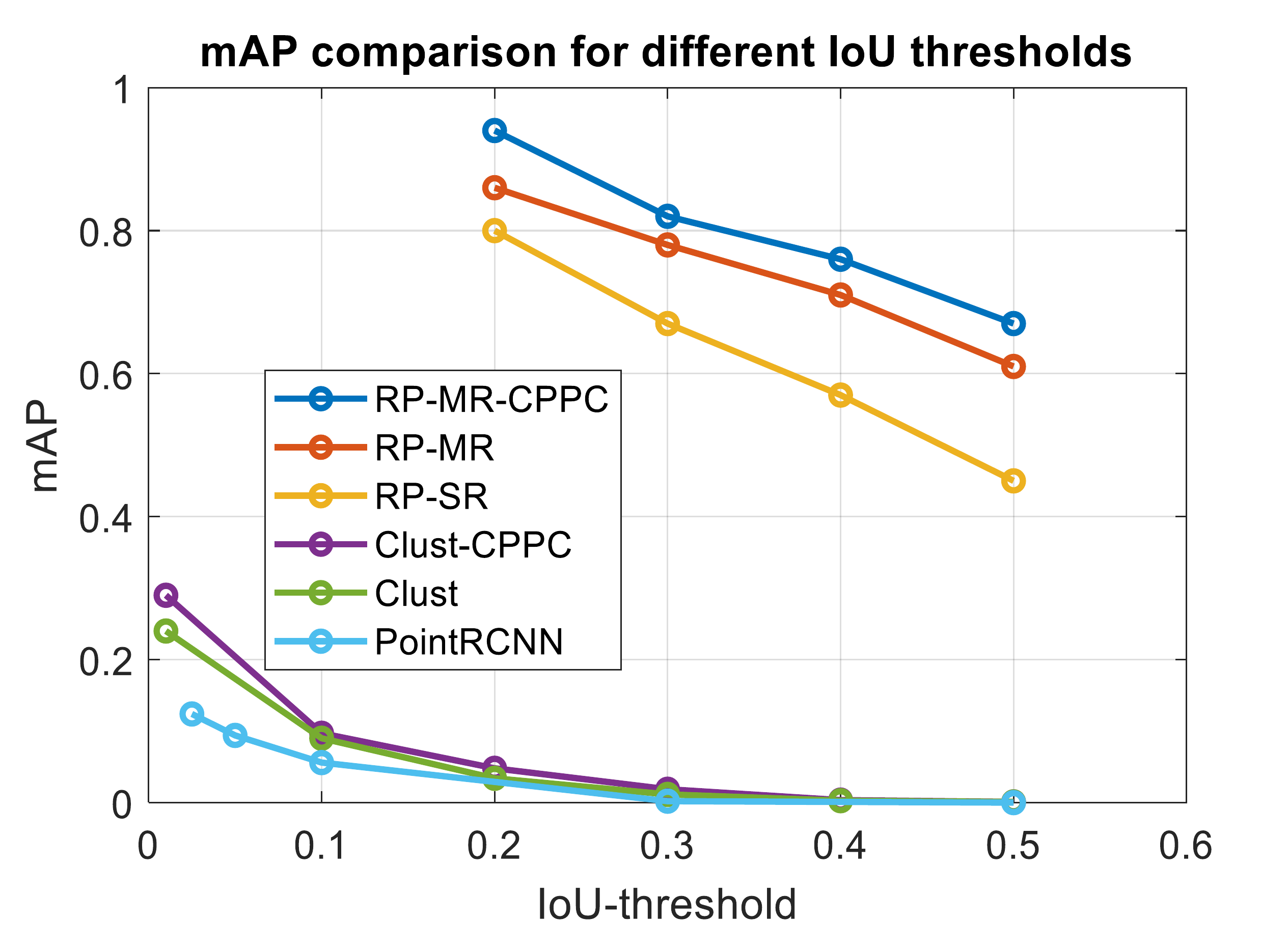}
 \vspace{-0.2in}
 \subcaption{}
\label{fig:singvmult}
\end{minipage}\quad
\begin{minipage}{0.3\linewidth}
 \includegraphics[width=\linewidth]{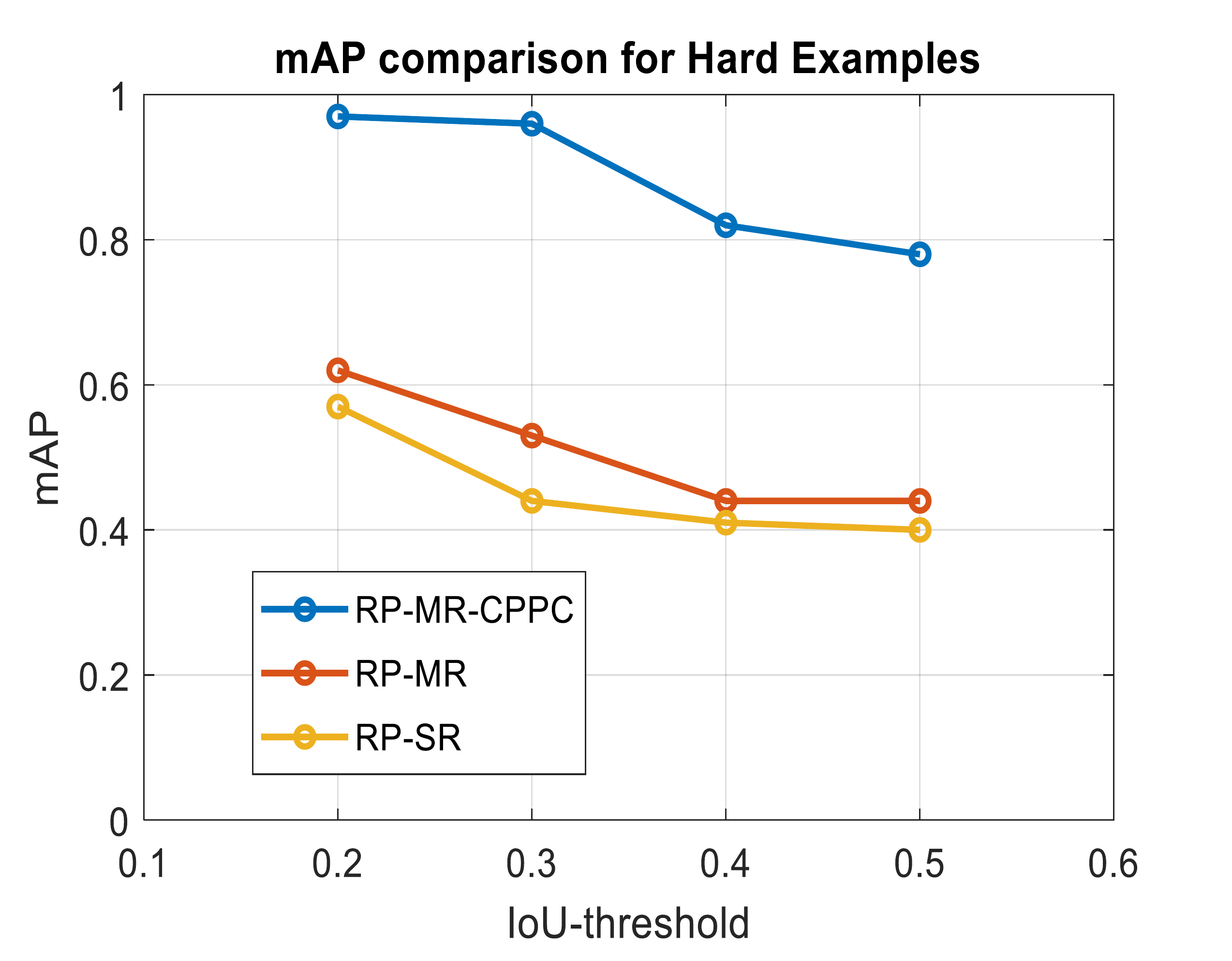}
 \vspace{-0.2in}
 \subcaption{}
 \label{fig:singlevmulthard}
 \end{minipage}
\begin{minipage}{0.3\linewidth}
 \includegraphics[width=\linewidth]{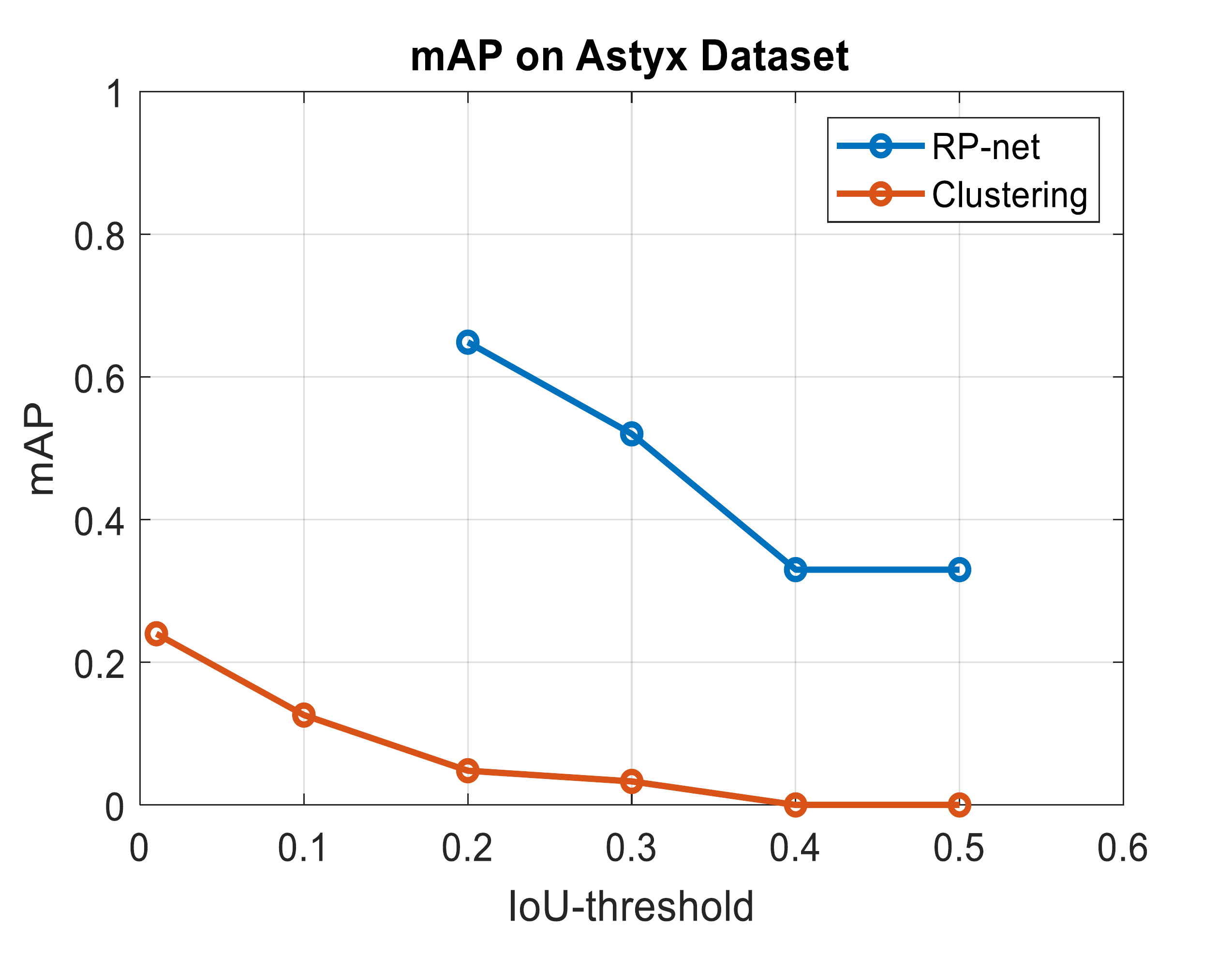}
 \vspace{-0.2in}
 \subcaption{}
 \label{fig:astyx}
 \end{minipage}
 \vspace{-0.15in}
\caption{a) \network outperforms PointRCNN and clustering baselines for bounding box estimation. RP-MR performs better than the RP-SR system. CPPC based combination further improves performance (RP-MR-CPPC). b) CPPC based fusion performs significantly better in hard examples due to noise estimation and prior heading direction using space-time coherence. c) mAP performance of \network on Astyx dataset compared to the clustering baseline.}
\label{fig:mAP_figures}
\end{figure*}


 In this section, we would comprehensively evaluate each part of \name. We would compare our results against the commonly used deterministic clustering~\cite{clustering} and deep learning based approaches~\cite{shi2018pointrcnn} for 3D bounding box estimation on point clouds. Figure~\ref{fig:qualitative} shows examples of predicted 3D bounding boxes by \name.

\textbf{Testing and Training data-set:} We test our system on the dataset collected by us (section 6) that contains 54000 radar frames. We use a train test split of 9:1. The data used for testing comes from separate data collection runs than training data to ensure the generalization of our approach. We also use the publicly available astyx dataset~\cite{astyx} for automotive radars to demonstrate the generalization of our approach. We also compare the performance of \name against LiDAR in bad weather conditions.

Following are the metrics we use to evaluate our system:
\begin{itemize}[leftmargin=*]
    \item \textbf{IoU}: IoU (Jaccard Index) is a measure of the overlap between the predicted bounding box and the ground truth box. 3D IoU is given by $\frac{\text{Intersection Volume}}{\text{Union Volume}}$. 2D IoU is defined for the top view  (also called Bird-eye-view (BEV)) rectangles of 3D bounding boxes, as $\frac{\text{Intersection Area}}{\text{Union Area}}$. Two equal-sized boxes with half overlap would have an IoU of 0.33. Hence, even an IoU of around 0.5 is generally regarded as a good overlap.
   
   \item \textbf{mAP}: (mean Average Precision) mAP is the area under the precision-recall (PR) curve, which is a measure of the number of actual boxes detected (recall) along with the accuracy of detections (precision). Specifically, precision is obtained for incremental recall values to get PR curve. 
    \begin{equation*}
       \begin{split}
        Precision &= TP/(TP+FP) \\
        Recall &= TP/(TP+FN)\\
        mAP &= Area(\text{precision-recall curve})   
       \end{split}
    \end{equation*}
    \vspace{-0.05cm}
    An estimation is regarded as a true positive (TP) if it is above a particular IoU (Intersection Over Union) threshold. Note that a higher recall rate can easily be obtained by predicting a large number of boxes, but at the cost of sacrificing precision (more False Positives (FP)) and vice-versa. A higher mAP means better performance on both accuracy (precision) and exhaustiveness (recall) of estimation. An FP could also be obtained because of noise. An FP generated due to noise will have a very small (almost 0) IoU with any ground truth box. Hence, in a lower IoU threshold regime, the mAP is more sensitive to the amount of noise and allows us to better compare our noise suppression performance.
    \vspace{-0.1cm}
\end{itemize}

In summary, \name achieves a median error of less than \textbf{37cm} in localizing the center of an object bounding box and a median error of less than \textbf{25cm} in estimating the dimensions of the bounding boxes. We use 2D IoU (BEV IoU) as the thresholds for our mAP metric \cite{Geiger2012CVPR}. \name achieves an mAP score of \textbf{0.67} for an IoU threshold of 0.5 and a score of \textbf{0.94} for a lower IoU threshold of 0.2, which is a \textbf{45\%} improvement over a single radar system. We further show that \network is easily generalizable to other datasets by evaluating its performance on \textit{Astyx} dataset~\cite{astyx} and achieve an mAP of \textbf{0.65} with IoU threshold of 0.2.

\vspace{-0.2cm}

\subsection{Performance on Bounding Box estimation}
Our entire system (dubbed RP-MR-CPPC) consists of multi-radar (MR) fusion to create \pcname (CPPC) and \network to estimate 3D bounding box. To individually compare the mAP performance of CPPC and \network we define the following baselines:
\begin{itemize}[leftmargin=*]
    \item \textbf{RP-SR:} \network on single radar data.
    \item \textbf{RP-MR}: \network on multiple radar data without \pcname fusion. The point clouds from multiple radars are simply added in the global coordinate system.
    \item \textbf{Clust}: A clustering based bounding box estimation baseline. A predefined size bounding box is estimated for each cluster found using DBSCAN, coupled with angle estimation using \textit{Principle Component Analysis} \cite{pca}
    \item \textbf{Clust-CPPC}: The clustering based approach used on \pcname. 
    \item \textbf{PointRCNN}: Official implementation of well-known LiDAR based 3D bounding box estimation network PointRCNN~\cite{shi2018pointrcnn} used on our collected dataset.
\end{itemize}
\vspace{-0.15cm}

Figure~\ref{fig:singvmult} shows the overall performance of these systems. The X-axis is the IoU thresholds used for the mAP values. Further, to best examine the performance improvement brought in by CPPC, we also evaluate the performance on a subset of validation set that contains hard examples following KiTTi evaluation framework~\cite{Geiger2012CVPR}. Hard examples are characterized by point clouds containing more than one-fourth of the points coming from noise, and the cars are undertaking complex maneuvers (sharp turns). The results of hard examples are shown in Figure~\ref{fig:singlevmulthard}.

\begin{figure}[t]
\centering
\includegraphics[width=\linewidth,]{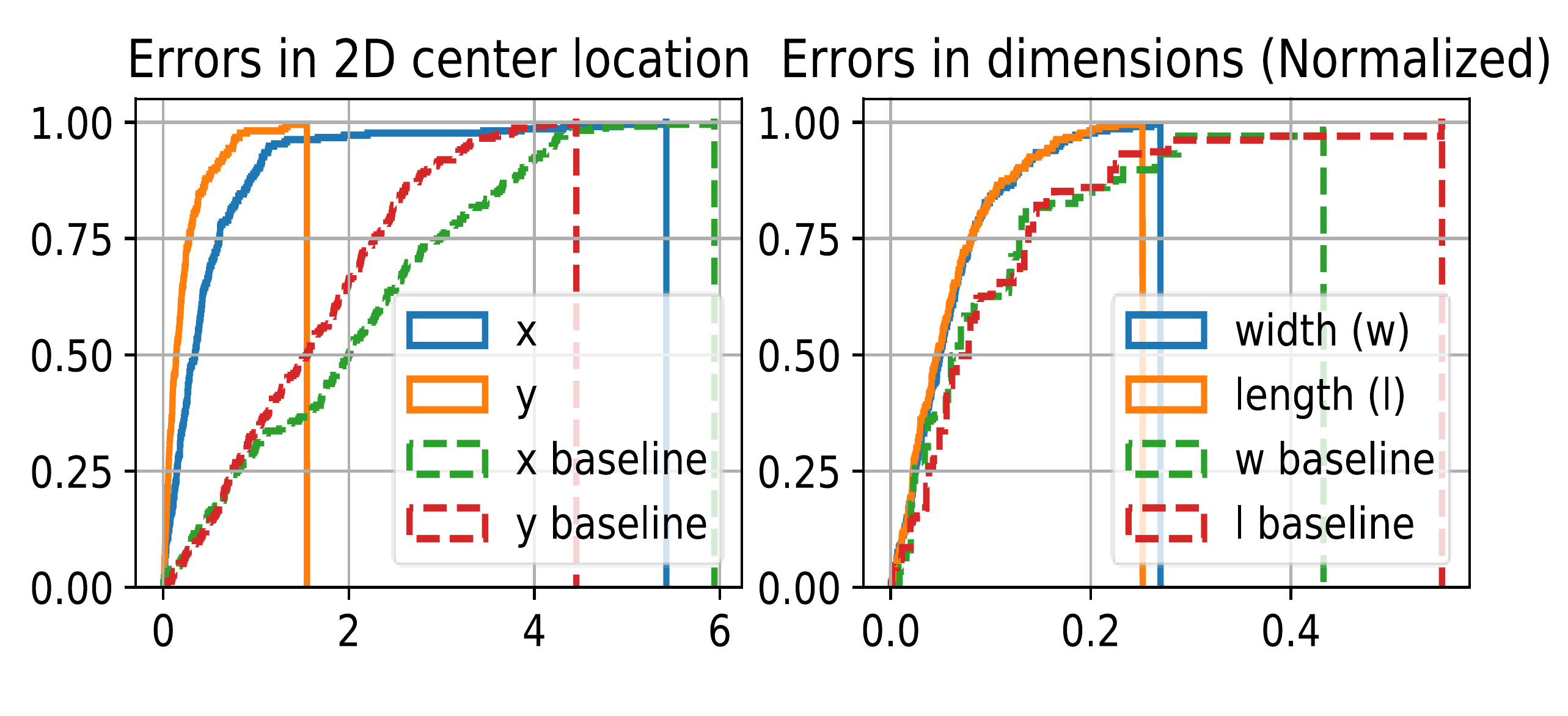}
\vspace{-0.6cm}
\caption{CDF plots for errors in center location and dimensions of predicted boxes (upwards is better). The x,y-coordinate is the depth and lateral dimensions, respectively. Compared to the clustering-based baseline, \network performs much better as it learns refinement from the data.}
\label{fig:cdf_dimensions}
\vspace{-0.7cm}
\end{figure}

\textbf{CDF comparison.} 
In this experiment, we evaluate the performance of \name on estimating the center coordinates of bounding boxes and its dimensions in 2D Bird-Eye-View (BEV). Proper estimation of bounding box parameters in BEV is critical for autonomous driving\cite{Geiger2012CVPR}. Figure \ref{fig:cdf_dimensions} shows the CDF plots of L1 absolute errors in center coordinates and dimensions of bounding boxes predicted by \network compared against the clustering-based baseline (clust-CPPC\cite{pca}). Errors in dimensions are normalized to actual ground truth box dimensions. Note that the clustering baseline does not have the refinement stage as in \network, and hence \name clearly outperforms the baseline. \network can learn the complex point cloud to bounding box mapping function (equation \ref{eq:detector_eqn_cppc}) using data and hence provides accurate bounding box estimation results.

\textbf{Performance of \network.}
Figure~\ref{fig:singvmult} \& \ref{fig:singlevmulthard} shows that \network performs significantly better compared to LiDAR-based network (PointRCNN) and the clustering baseline. This result is expected as a LiDAR-based network can not be directly applied to radar point clouds due to inherent differences in point cloud distribution. LiDAR-based networks are designed to work in cases where background points are much larger compared to the foreground(points on objects of interest). Also, the poor performance of the clustering baseline conveys the complexity of the task and the requirement of data based learning. The performance for the clustering baseline increases with lower threshold values because it lacks any refinement of the bounding box to get the correct dimensions of the box. \network architecture is specifically designed to handle the sparsity and non-uniformity of radar point clouds by learning to model their intricacies and using a special set of anchor boxes. 

\textbf{Effect of using multiple radars.}
Multiple radar fusion based approach (RP-MR) outperforms the single radar baseline (RP-SR) in all the cases (Figure~\ref{fig:singvmult} \& \ref{fig:singlevmulthard} ). 
The mAP value increases from 0.45 to 0.61 for IoU threshold 0.5 as we move to multiple radars from a single radar. The reason behind this is the specularity of mmWaves that prevent all the features from being captured and hence makes it hard to get the dimensions of the box right, inherently affecting the mAP values. Using multiple radars decreases the adverse effect of specularity, as evident from the increase in mAP values from single to multiple radar approaches.

\textbf{Performance with multiple cars}
We evaluate the performance of different networks in detecting the vehicles present in the scene. Table \ref{tab:recall} shows the recall rate achieved with the different number of cars present in the scene. The performance drops when the number of vehicles in the scene increases due to occlusions. Still, RP-MR-CPPC outperforms the single radar non-CPPC based models in all the cases, which proves the superiority of our algorithm over baselines. 
\begin{table}[h!]

\begin{tabular}{|c|c|c|c|c|c|}
\hline
\textit{Model vs \# Cars} & \textbf{1}    & \textbf{2}    & \textbf{3}    & \textbf{4}    & \textbf{Entire Dataset} \\ \hline
\textbf{RP-SR}     & 0.4           & 0.31          & 0.2           & 0.1           & 0.32                    \\
\textbf{RP-MR}      & 0.61          & 0.45          & 0.3           & 0.28          & 0.47                    \\
\textbf{RP-MR-CPPC}          & \textbf{0.75} & \textbf{0.52} & \textbf{0.41} & \textbf{0.38} & \textbf{0.6}            \\ \hline
\end{tabular}
\caption{Recall rate for different number of cars in the scene.}
\label{tab:recall}
\vspace{-0.8cm}
\end{table}

\begin{table}[h!]
\begin{tabular}{|c|c|c|c|c|}
\hline

\textbf{Model}                & \textbf{Velocity} & \textbf{Intensity} & \textbf{CPPC} & \textbf{mAP }                   \\ \hline
\multirow{3}{*}{Ablated Model} & \xmark      & \xmark   & \xmark      & 0.56 \\ 
                        &   \cmark      & \xmark    & \xmark      & 0.60\\ 
                       &      \cmark    &  \cmark    & \xmark     & 0.61  \\ \hline
Final Model              &      \cmark    & \cmark   &     \cmark    & \textbf{0.67}      
            \\ \hline

\end{tabular}
\caption{Ablation studies: change in performance with different input channels to the network.}
\label{tab:mAP}
\vspace{-0.8cm}
\end{table}

\textbf{Effect of using CPPC.}
Figure~\ref{fig:singvmult} shows that using a CPPC based fusion approach is better than just a naive combination of multiple radar point clouds (RP-MR). In the hard examples (figure~\ref{fig:singlevmulthard}) the effect is even more significant. The performance increases from \textbf{0.44} to \textbf{0.78} by using CPPC on multiple radars for IoU threshold of 0.5. Note that even for the clustering based baseline, using CPPC fusion improves the results for bounding box estimation. The space-time coherence in CPPC reduces the noise, thereby minimizing false positives and improving results.

\textbf{Comparison to other approaches (LiDAR/sensor fusion).} In order to put these results into perspective, we provide some results from LiDAR-based approaches. PointRCNN~\cite{shi2018pointrcnn} achieves 0.78 mAP on KiTTi LiDAR dataset~\cite{Geiger2012CVPR}. A camera and radar fusion approach, on a dataset provided by Astyx GMBH~\cite{astyx}, achieves an mAP value of 0.45 on the complete dataset. Although it is not fair to directly compare with the LiDAR results, as the underlying datasets and IoU thresholds used are different (LiDAR-based approaches use a higher IoU threshold of 0.7), but we want to emphasise on the fact that it is possible to achieve LiDAR or camera like perception using just radars. Moreover, a higher resolution radar can produce a denser point cloud, that would further increase the IoU performance using our method.

\begin{table}[h!]

\begin{tabular}{|c|c|c|c|}
\hline
\textbf{Model}       & \textbf{mAP (IoU>0.5)} & \textbf{Time}\\ \hline
256 channels &       0.50     & 0.018 seconds (55 fps)\\ 
1024 channels  &       0.68  &  0.0208 seconds (48 fps) \\ \hline
\end{tabular}
\caption{Effect of model size on time and performance}
\label{tab:modelvalidation}
\vspace{-0.8cm}
\end{table}

\vspace{-0.3cm}
\subsection{Model Generalization on Astyx dataset}
In order to evaluate the generalization of our network, we test our network on the recently released Astyx dataset~\cite{astyx}. The released data set contains only 545 scene point clouds, which are quite less in order to allow efficient training. Also, the dataset contains only a single radar, which prevents us from applying our entire processing chain to the dataset. We divide the dataset into 5\% test (and rest training) set and evaluate the object detection performance. \network is able to achieve a performance of \textbf{0.65} mAP on this dataset with an IoU threshold of 0.2 while the clustering baseline achieves a significantly lower mAP of 0.05, as shown in  Figure~\ref{fig:astyx}.

\vspace{-0.2cm}

\subsection{Effect of input channels}

In this experiment, we would perform the ablation study for adding different channels to our input. We report the mAP score changes with the incorporation of a different number of input channels. Table~\ref{tab:mAP} summarizes the performance of this experiment. As expected, adding the velocity as an input channel brings the performance improvement as it provides the context about the direction of the vehicle's movement. Adding the intensity values do not have much effect on the performance. We hypothesize that as the target objects are only cars, there isn't much difference between the intensity of reflections. Lastly, using \pcname based fusion further improves the performance to mAP score of 0.67 because of space-time coherence.

\subsection{Model Architecture validation}
In this section, we evaluate the effect of model size on the performance. We change the number of channels for the representative feature vector of anchor boxes and compare the performance. Table~\ref{tab:modelvalidation} shows the result of our analysis. This experiment sheds light on the trade-off of time and accuracy. Clearly, with the increase in the number of layers from 256 to 1024, the model performs better (0.18 increase in mAP ) but lags on time (0.01 sec slower).

\begin{figure}[t]
\centering
\includegraphics[width=0.99\linewidth,]{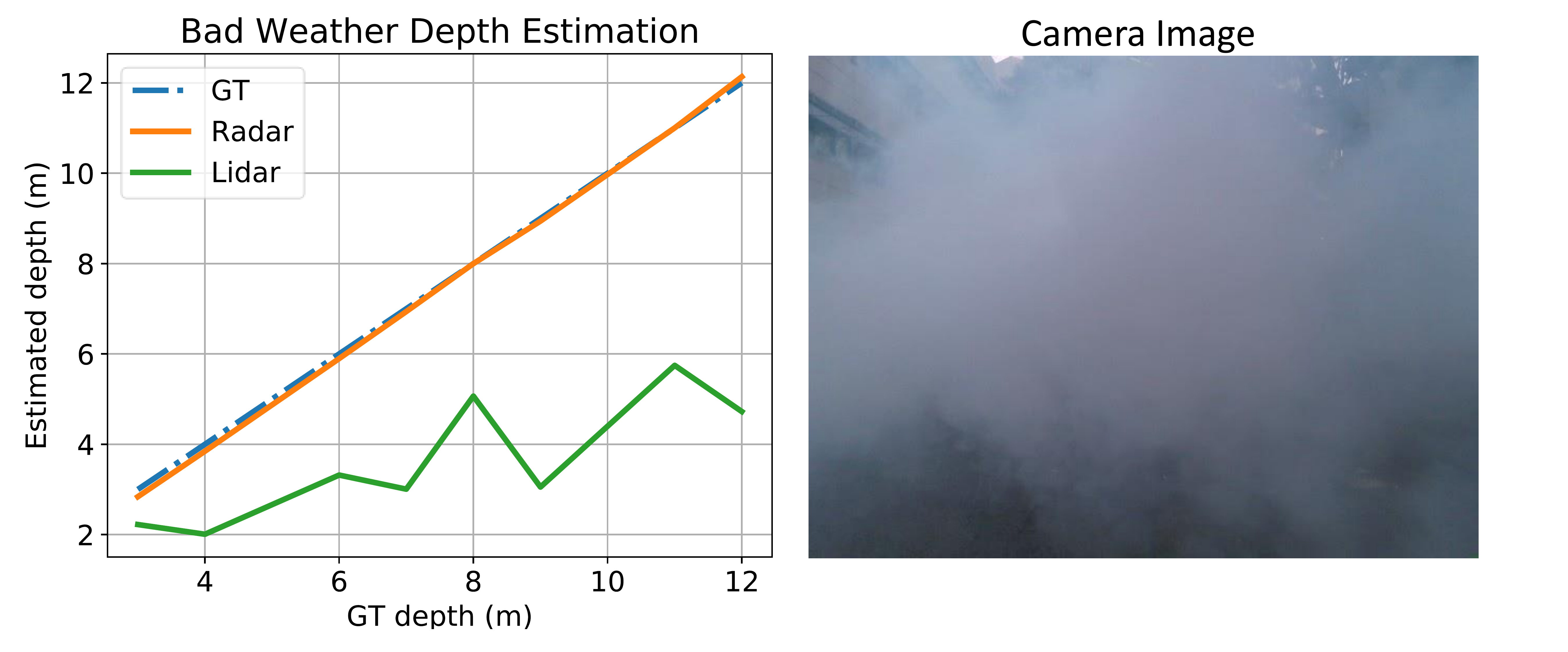} 
\caption{Performance comparison of different sensors in the presence of adverse conditions. The left plot shows the depth estimation performance of Radar and LiDAR for an object directly in front of the sensor in the presence of fog. The right figure shows the camera image for the experiment.}
\label{fig:badweather}
\vspace{-0.24in}
\end{figure}
One of the major requirements of an autonomous perception system is real-time performance. Our best model makes inference at 50fps while running on an NVIDIA GTX 1080 Ti GPU(Table~\ref{tab:modelvalidation}). Normal human reaction time is 100ms, which means that our model can provide 5 different scene perception outcomes until a normal human reacts to an event. Thus, \name has huge potential to be deployed on a real-time autonomous vehicular system.

\vspace{-0.1cm}
\subsection{Performance in bad weather}
The main objective of using radar as a primary sensory modality is to enable all-weather perception. To compare the performance of different sensing modalities in adverse weather scenarios, we perform experiments in foggy conditions. We use an artificial fog generator to simulate adverse weather conditions and estimate the performance of the depth estimation of a single-vehicle present in front of the sensor. The first point of return is regarded as the predicted depth. Figure~\ref{fig:badweather} shows the performance comparison between LiDAR and Radar (the results are an averaged over 100 frames).
LiDAR and camera get severely affected in the presence of fog, which would pose severe problems for tasks like depth estimation. Radar, on the other hand, remains unaffected. Moreover, past work \cite{daniel2017low,kutila2018automotive} conduct large scale experiments using fog chambers and artificial rain generators to show the prominence of radar sensors compared to LiDAR sensors in adverse weather conditions.


%% file: 9-related.tex
\section{Related work}
\label{sec:related}

The techniques designed in \name are fundamental and could be used to improve LiDAR, ultra-sonic, and infra-red imaging other radar applications. Spatial separation analysis could be used for other applications to identify optimal separation. Similarly, \pcname and RP-net can potentially be extended to benefit all the radar sensing applications relying on radar point clouds. Our work is closely related to the work on the following: 

\noindent\textbf{MIMO radar and Multi-radar fusion.}
Radars have been long-standing sensors in the field of surveillance and detection. Past works have considered the fusion of multiple radar sensors for aerial surveillance~\cite{favalli1996multi,xu2017multi,jouny2007target,petsios2008solving} but do not have much relevance to self-driving scenarios involving high-resolution MIMO radars. In \cite{petsios2008solving}, a nearest neighbor based association scheme is given, which is again limited to aerial surveillance. Also, the presence of ghost objects is not taken into account, which is a major part of real-world automotive radar data. We perform all our analysis on real data of dynamic traffic environments. A multi radar setup is considered in ~\cite{scheel2016multi} with a common field of view. They use \textit{Labelled Multi Burnoulli} filter to track and associate the detections in multiple radars but fail to leverage the multi-sensor input or optimize spatial separation. They instead treat it as a single sensor. A multi radar system deployed in ~\cite{metzner2018multi}, compares performance with a single radar system for the task of blind-spot detection using Kalman filter based tracking. They use multiple radars to achieve a wider FoV by simply combining the detections generated from two radars and could not report any significant performance gains. \name takes advantage of multiple radars for increasing the point cloud density while minimizing noise recorded in the scene, as opposed to getting a wider FoV.

\noindent\textbf{Synthetic Aperture Radar}
Synthetic Aperture Radar(SAR) \cite{fembacher2018real,gisder2019synthetic,gishkori2019imaging} based techniques are used to increase the resolution of radar, but they have limitations. SAR approaches can not resolve the specularity problem of mmWaves, which requires a different viewpoint as provided by multiple radars in \name. For a vehicle moving on the road, a SAR based approach needs to be implemented using the motion of the car as done in \cite{fembacher2018real,gisder2019synthetic}, instead of physically moving the radar. This approach to SAR only increases the resolution on the sides of the ego vehicle (side-looking SAR). Applications of this method include parking spot detection, but object detection in front of vehicles is not improved. A forward-looking SAR is implemented in \cite{gishkori2019imaging} by using a beam scanning 300 GHz radar, which is different from the 77 GHz MIMO radar we used. This approach tries to improve angular resolution at small boresight angles. It does not take care of noise generated by the multipath and random clutter, which is handled by \name's CPPC formulation. \name’s approach for multi-radar fusion generalizes to the fundamental distribution of radar point clouds and could be applied to all the above work and reap significant benefits for tackling specularity, noise cancellation, and sparsity.



\noindent \textbf{Deep Learning based 3D bounding box.}
With the abundance of open-source LiDAR datasets \cite{Geiger2012CVPR,nuscenes2019,waymo_open_dataset} for autonomous driving, object detection on LiDAR point clouds has seen numerous advancements. Point clouds are converted into voxel grids for object detection with bounding box prediction in \cite{maturanavoxnet,Qi_2016}. Direct object detection on point cloud data is performed in \cite{Qi_2018,wang2019frustum}. However, they use calibrated 2D RGB images to segment instance-based frustums from a point cloud. A camera-radar fusion approach for bounding boxes is provided in \cite{astyx_deep} but would be infeasible for all-weather perception tasks. PointRCNN \cite{shi2018pointrcnn} performs 3D bounding box estimation only using LiDAR point clouds. However, due to the sparsity of radar point clouds, the approach fails to capture enough context for segmentation. \network (Our approach) combines the region proposal and segmentation to overcome the sparsity problem. Andreas et al. An approach based on frustum pointnet is followed in \cite{danzer20192d}, where patches are considered throughout the 2D image followed by a segmentation head that identifies the radar targets lying on the object of interest. However, the patch proposal can not be readily extended to 3D detection.


\noindent \textbf{\vspace{-2pt}Radar sensing for health-care and indoors.} 
FMCW is a popular technique for RF sensing. FMCW based radars (WiFI frequency) have also been used for indoor health monitoring and localization applications \cite{zhao2016emotion, adib2015smart, adib2015multi, hsu2019enabling, yue2018extracting, tian2018rf, hsu2017zero, hsu2017extracting}. Low-frequency radars (WiFI frequency) have been used \cite{zhao2018through, zhao2018rf, zhao2017learning, tian2018rf} for tasks like human pose estimation, fall detection, and sleep stage detection using deep learning. All these works use low-frequency FMCW based and can not be readily applied in outdoor settings. mmWave radars have been used indoors for ego-motion estimation and corridor mapping \cite{lu2020see,almalioglu2020milli}. The techniques cross-potential point cloud and \network can be used by the indoor radars as well, as they are complementary to all existing work and provides a neat framework that is inspired by the distribution of the radar point clouds. For example, reflection tracking is a key technique for all the above work. One could apply \name's cross-potential based ideas to the reflections and their tracking to make them robust against noise and dynamic multi-path~\cite{adib2015multi}.

\noindent \textbf{\vspace{-2pt}Automotive Radar data processing using deep learning.}
Deep learning has been applied to various stages of radar data processing. Past works \cite{Visentin2017ClassificationOO,lombacher2016potential} perform object classification on radar data using CNNs. Pointnet++~\cite{qi2017pointnet} is used directly on radar point clouds to perform semantic segmentation in \cite{Schumann2018SemanticSO}. These approaches do not consider the major challenges of specularity and sparsity with radar point clouds. A GAN based approach\cite{haitham} has been proposed to image cars using synthetic aperture based sensing but is limited to static scenes. 2D box prediction using radar tensors for the highway has been done in \cite{bencemajor}. Their approach is only limited to traffic scene where vehicles are moving straight and can not retrieve the 3D pose of vehicles. The special distribution of radar point clouds requires an approach specifically designed for radar data, such as CPPC. We believe \pcname can help existing work to improve the performance significantly.

\vspace{-0.3cm}


%% file: 910-discussion.tex
\section{Limitations and Future Work}


Pointillism presents a system that enables object detection using only automotive radars. It tackles the fundamental problems of noise, specularity, and sparsity present in radar point clouds by using multiple radars at once. The radars used during the evaluation of our system offer an angular resolution of 15 degrees. Current high-resolution automotive radars provide an angular resolution of 1 degree that would potentially lead to better bounding box refinement and hence better IoU values. This means that the current mAP performance of pointillism at higher IoU thresholds (section \ref{sec:evaluation}) could further be improved by using higher-resolution radars. 
A natural extension of this work is the time-based tracking of dynamic objects present in the scene. In section 4, we describe how we use space coherence to eliminate noise and time coherence to get a pose estimate. A tracking based formulation could possibly combine the noise filtering with pose estimation by looking at time-based data from multiple radars. Secondly, the radars also provide doppler estimates of the objects. We use doppler as an input channel to RP-net. However, it could also be used to classify different types of road users like pedestrians, bikes, and cars. The concepts introduced in our paper relate to the fundamental nature of radar point clouds and can be extended to enable more complicated tasks than object detection.